\documentclass[10pt,twocolumn,letterpaper]{article}

\usepackage{cvpr}      

\usepackage{graphicx}
\usepackage{amsmath}
\usepackage{amssymb}
\usepackage{booktabs}
\usepackage{multirow}
\usepackage{array}
\usepackage[table]{xcolor}
\usepackage{xcolor}
\usepackage[font=small]{caption}
\usepackage[pagebackref,breaklinks,colorlinks]{hyperref}

\usepackage[capitalize]{cleveref}
\crefname{section}{Sec.}{Secs.}
\Crefname{section}{Section}{Sections}
\Crefname{table}{Table}{Tables}
\crefname{table}{Tab.}{Tabs.}
\definecolor{lightyellow}{RGB}{182, 215, 242}
\definecolor{lightgray}{RGB}{251,  227, 213}

\newcommand{\bestcell}{\cellcolor{lightyellow}}
\newcommand{\secondcell}{\cellcolor{lightgray}}

\newcolumntype{P}[1]{>{\centering\arraybackslash}p{#1}}

\def\cvprPaperID{1126} 
\def\confName{CVPR}
\def\confYear{2023}

\begin{document}

\title{HexPlane: A Fast Representation for Dynamic Scenes}

\author{
    Ang Cao \hspace{4mm}
    Justin Johnson \\*[3mm]
    University of Michigan, Ann Arbor \\
     {\tt \small \{ancao, justincj\}@umich.edu}
}
\maketitle

\begin{abstract}
    Modeling and re-rendering dynamic 3D scenes is a challenging task in 3D vision.  
    Prior approaches build on NeRF and rely on implicit representations.
    This is slow since it requires many MLP evaluations, constraining real-world applications.
    We show that dynamic 3D scenes can be explicitly represented by six planes of learned features, leading to an elegant solution we call HexPlane.
    A HexPlane computes features for points in spacetime by fusing vectors extracted from each plane, which is highly efficient.
    Pairing a HexPlane with a tiny MLP to regress output colors and training via volume rendering gives impressive results for novel view synthesis on dynamic scenes, matching the image quality of prior work but reducing training time by more than $100\times$.
    Extensive ablations confirm our HexPlane design and show that it is robust to different feature fusion mechanisms, coordinate systems, and decoding mechanisms.
    HexPlane is a simple and effective solution for representing 4D volumes, and we hope they can broadly contribute to modeling spacetime for dynamic 3D scenes.\footnote{Project page: \url{https://caoang327.github.io/HexPlane}.}
\end{abstract}



\section{Introduction}
Reconstructing and re-rendering 3D scenes from a set of 2D images is a core vision problem which can enable many AR/VR applications.
The last few years have seen tremendous progress in reconstructing \emph{static} scenes,
but this assumption is restrictive: the real world is \emph{dynamic}, and in complex scenes motion is the norm, not the exception.

Many current approaches for representing dynamic 3D scenes rely on \emph{implicit} representations, building on NeRF~\cite{mildenhall2020nerf}.
They train a large multi-layer perceptron (MLP) that inputs the position of a point in space and time, and outputs either the color of the point~\cite{li2021neural, li2022neural} or a \emph{deformation} to a canonical static scene~\cite{gao2021dynamic,pumarola2021d,park2021hypernerf,park2021nerfies}.
In either case, rendering images from novel views is expensive since each generated pixel requires many MLP evaluations.
Training is similarly slow, requiring up to days of GPU time to model a single dynamic scene;
this computational bottleneck prevents these methods from being widely applied.

\begin{figure}[t!]
	\centering
    {\includegraphics[width=0.5\textwidth]{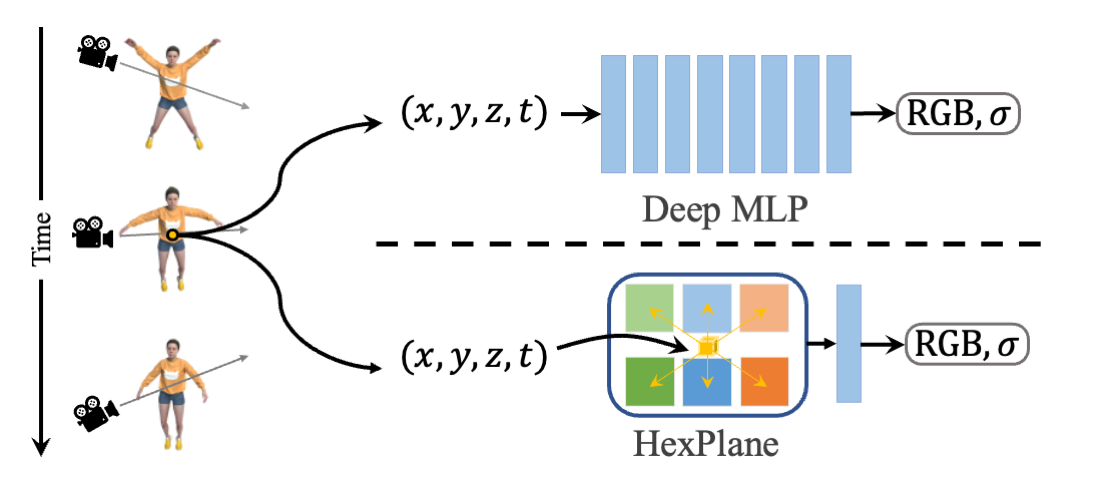}}
	\captionof{figure}{\textbf{HexPlane for Dynamic 3D Scenes.}
	Instead of regressing colors and opacities from a deep MLP, we explicitly compute features for points in spacetime via HexPlane.
	Pairing with a tiny MLP, it allows 
	above $100\times$ speedups with matching quality.}
	\label{fig:teaser}
	\vspace{-6mm}
\end{figure}

Several recent methods for modeling \emph{static} scenes have demonstrated tremendous speedups over NeRF through the use of \emph{explicit} and \emph{hybrid} methods~\cite{yu2021plenoctrees,chen2022tensorf,sun2022direct,mueller2022instant}.
These methods use an explicit spatial data structure that stores explicit scene data~\cite{yu2021plenoctrees,fridovich2022plenoxels} or features that are decoded by a tiny MLP~\cite{chen2022tensorf,sun2022direct,mueller2022instant}.
This decouples a model's \emph{capacity} from its \emph{speed}, and allows high-quality images to be rendered in realtime~\cite{mueller2022instant}.
While effective, these methods have thus far been applied only to static scenes.

In this paper, we aim to design an explicit representation of dynamic 3D scenes, building on similar advances for static scenes.
To this end, we design a \emph{spatial-temporal} data structure that stores scene data.
It must overcome two key technical challenges.
First is \emph{memory usage}.
We must model all points in both space and time; na\"ively storing data in a dense 4D grid would scale with the fourth power of grid resolution which is infeasible for large scenes or long durations.
Second is \emph{sparse observations}.
Moving a single camera through a static scene can give views that densely cover the scene;
in contrast, moving a camera through a dynamic scene gives just one view per timestep.
Treating timesteps independently may give insufficient scene coverage for high-quality reconstruction,
so we must instead share information across timesteps.

We overcome these challenges with our novel HexPlane architecture.
Inspired by factored representations for static scenes~\cite{peng2020convolutional,chen2022tensorf,chan2022efficient},
a HexPlane decomposes a 4D spacetime grid into six \emph{feature planes} spanning each pair of coordinate axes
(\eg $XY$, $ZT$).
A HexPlane computes a feature vector for a 4D point in spacetime by projecting the point onto each feature plane,
then aggregating the six resulting feature vectors.
The fused feature vector is then passed to a tiny MLP which predicts the color of the point;
novel views can then be rendered via volume rendering~\cite{mildenhall2020nerf}.

Despite its simplicity, a HexPlane provides an elegant solution to the challenges identified above.
Due to its factored representation, a HexPlane's memory footprint only scales quadratically with scene resolution.
Furthermore, each plane's resolution can be tuned independently to account for scenes requiring variable capacity in space and time.
Since some planes rely only on \emph{spatial} coordinates (\eg $XY$), by construction a HexPlane encourages sharing information across disjoint timesteps.

Our experiments demonstrate that HexPlane is an effective and highly efficient method for novel view synthesis in dynamic scenes.
On the challenging Plenoptic Video dataset~\cite{li2022neural} we match the image quality of prior work but improve training time by ${>}100\times$;
we also outperform prior approaches on a monocular video dataset~\cite{pumarola2021d}.
Extensive ablations validate our HexPlane design and demonstrate that it is robust to different feature fusion mechanisms, coordinate systems (rectangular vs. spherical), and decoding mechanisms (spherical harmonics vs. MLP).

HexPlane is a simple, explicit, and general representation for dynamic scenes.
It makes minimal assumptions about the underlying scene, and does not rely on deformation fields or category-specific priors.
Besides improving and accelerating view synthesis, we hope HexPlane will be useful for a broad range of research in dynamic scenes~\cite{Singer2023TextTo4DDS}. 

\section{Related Work}
\par \noindent {\bf Neural Scene Representations.}
Using neural networks to implicitly represent 3D scenes~\cite{mescheder2019occupancy, niemeyer2020differentiable, sitzmann2019deepvoxels, wang2021ibrnet, Sun2022Neural3R,Sitzmann2019SceneRN} has achieved exciting progress recently.
NeRF~\cite{mildenhall2020nerf} and its variants~\cite{yen2022nerfsupervision,Tancik2022BlockNeRFSL, zhang2020nerf++,barron2021mip,barron2022mip,niemeyer2022regnerf,verbin2021ref, mildenhall2022nerf} show impressive results on novel view synthesis~\cite{wang2021ibrnet,chen2021mvsnerf,yu2021pixelnerf, zhou2022sparsefusion} and many other applications 
including 3D reconstruction~\cite{Zhang2022NeRFusionFR,zhu2022nice, martin2021nerf, Sun2022Neural3R,Zhang2021NeRSNR}, semantic segmentation~\cite{qian2021recognizing,Zhi:etal:ICCV2021,kobayashi2022distilledfeaturefields}, generative model~\cite{chan2021pi,chan2022efficient, schwarz2020graf,Niemeyer2020GIRAFFE,Deng2022GRAMGR, xie2022high}, and 3D content creation~\cite{neumesh,poole2022dreamfusion,jain2021dreamfields,wang2022clip,Ouyang2022RealTimeNC,Zhang2022ARFAR,lin2022magic3d}.

Implicit neural representations exhibit remarkable rendering quality, but they suffer from slow rendering speeds due to the numerous costly MLP evaluations required for each pixel. 
To address this challenge, many recent papers propose \emph{hybrid} representations that combine a fast explicit scene representation with learnable neural network components, providing significant speedups over purely implicit methods.
Various explicit representations have been investigated, including sparse voxels~\cite{sun2022direct,fridovich2022plenoxels, schwarzvoxgraf, liu2020neural}, low-rank components~\cite{chan2022efficient,peng2020convolutional,chen2022tensorf, Lionar2020DynamicPC}, point clouds~\cite{xu2022point, Zuo2022ViewSW, huang2022ponder, zheng2022pointavatar,cao2022fwd} and others~\cite{mueller2022instant, Takikawa2022VariableBN, Lombardi2021MixtureOV, zhang2023efficient, chen2023factor}.
However, these approaches assume static 3D scenes, leaving explicit representations for dynamic scenes unexplored.
This paper provides an explicit model for dynamic scenes, substantially accelerating prior methods that rely on fully implicit methods.

\par \noindent {\bf Neural Rendering for Dynamic Scenes.}
Representing dynamic scenes by neural radiance fields is an essential extension of NeRF, enabling numerous real-world applications~\cite{Zhao2021HumanNeRFGN, Peng2021NeuralBI, Su2021ANeRFAN, Li2022TAVATA, Xu2021HNeRFNR, ost2021neural, Zhang2021EditableFV}.
One line of research represents dynamic scenes by extending NeRF with an additional time dimension~(T-NeRF) or additional latent code~\cite{gao2021dynamic, xian2021space, li2022neural, li2021neural}. 
Despite the ability to represent general typology changes, they suffer from a severely under-constrained problem, requiring additional supervision like depths, optical flows or dense observations for decent results.
Another line of research employs individual MLPs to represent a deformation field and a canonical field~\cite{pumarola2021d, park2021nerfies, park2021hypernerf, du2021neural, tretschk2021non, yuan2021star}, where the canonical field depicts a static scene, and the deformation field learns coordinate maps to the canonical space over time.
We propose a simple yet elegant solution for dynamic scene representation using six feature planes, making minimal assumptions about the underlying scene.

Recently, \emph{MAV3D}~\cite{Singer2023TextTo4DDS} adopted our design for text-to-4D dynamic scene generation, demonstrating an exciting direction for dynamic scenes beyond reconstruction.

\par \noindent {\bf Accelerating NeRFs.}
Many works have been proposed to accelerate NeRF at diverse stages. 
Some methods improve \emph{inference} speeds of trained NeRFs by optimizing the computation~\cite{Reiser2021KiloNeRFSU, garbin2021fastnerf, yu2021plenoctrees,hedman2021baking}.
Others reduce the \emph{training} times by learning a generalizable model~\cite{chen2021mvsnerf,Wang_2022_CVPR, wang2021ibrnet, johari2022geonerf}.
Recently, rendering speeds during \emph{both stages} are substantially reduced by using explicit-implicit representations~\cite{chen2022tensorf, fridovich2022plenoxels,sun2022direct,mueller2022instant,Liu2020NeuralSV,chan2022efficient}.
In line with this idea, we propose an explicit representation for dynamic fields to accelerate dynamic NeRFs.

Very recently, several concurrent works have aimed to accelerate dynamic NeRFs.
\cite{fang2022fast, Guo2022NeuralDV, Liu2022DeVRFFD, Gan2022V4DVF, Wang2022MixedNV} use \emph{time-aware} MLPs to regress spacetime points' colors or deformations from canonical spaces.
However, they remain partially implicit for dynamic fields, as they rely on MLPs with time input to obtain spacetime features.
In contrast, our paper proposes a more elegant and efficient explicit representation for dynamic fields without using time-aware MLPs. Like \cite{Li2022StreamingRF}, \emph{NeRFPlayer}\cite{Song2022NeRFPlayerAS} uses a highly compact 3D grid at each time step for 4D field representation ,
which results in substantial memory costs for lengthy videos.

\emph{Tensor4D}~\cite{Shao2022Tensor4DE} shares a similar idea as ours, which represents dynamic scenes with 9 planes and multiple MLPs. 
\emph{D-TensoRF}~\cite{jang2022d} regards dynamic fields as 5D tensors and applies CP/MM decomposition on them for compact representation. 
Our paper is most closely related to \emph{$K$-Planes}~\cite{fridovich2023k}, which also employs six feature planes for representation.

\begin{figure*}
    \centering
    \includegraphics[width=0.9\textwidth]{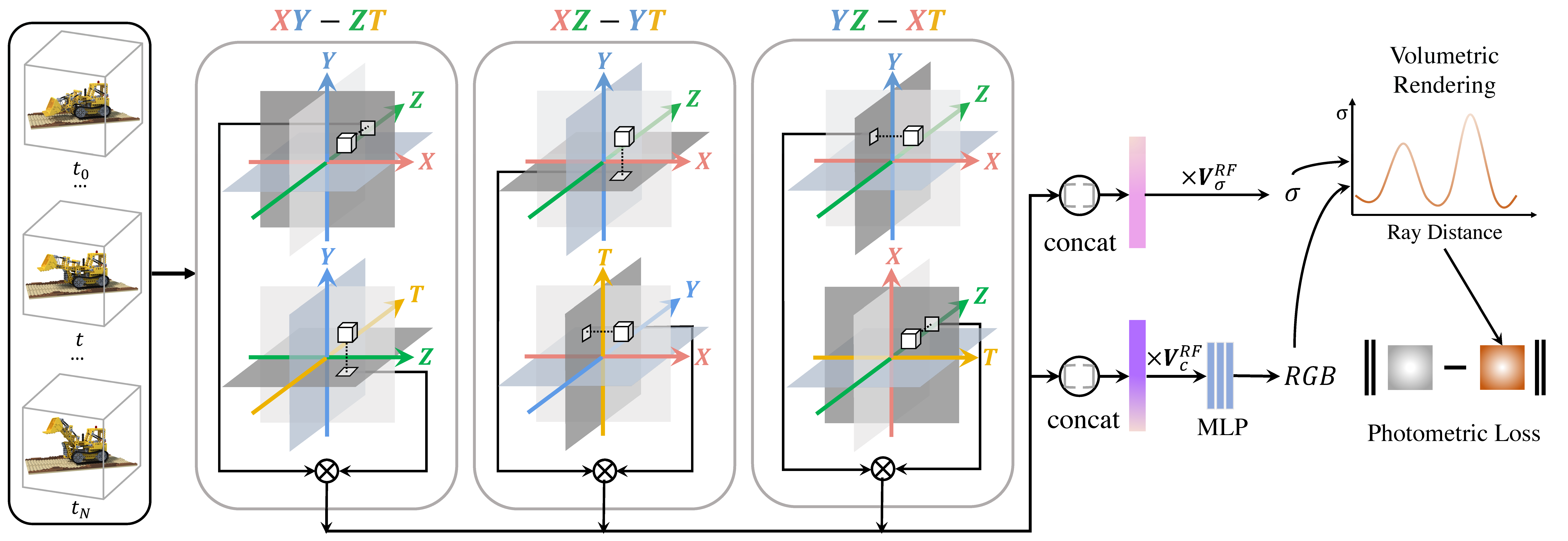}
    \vspace{-2mm}
	\captionof{figure}{{\bf Method Overview.}
    HexPlane contains six feature planes spanning each pair of coordinate
    axes (e.g. XY , ZT ).
To compute features of points in spacetime, it multiplies feature vectors extracted from paired planes and concatenated multiplied results into a single vector, which are then multiplied by $\mathbf{V}^{RF}$ for final results.
    RGB colors are regressed from point features using a tiny MLP and 
    images are synthesized via volumetric rendering.
    HexPlane and the MLP are trained by photometric loss between rendered and target images.} 
	\label{fig:model}
    \vspace{-4mm}
\end{figure*}

\section{Method}
Given a set of posed and timestamped images of a dynamic scene, we aim to fit a model to the scene that allows rendering new images at novel poses and times.
Like NeRF~\cite{mildenhall2020nerf}, a model gives color and opacity for points in spacetime; images are rendered via differentiable volumetric rendering along rays.
The model is trained using photometric loss between rendered and ground-truth images.

Our main contribution is a new \emph{explicit} representation for dynamic 3D scenes, which we combine with a small \emph{implicit} MLP to achieve novel view synthesis in dynamic scenes.
An input spacetime point is used to efficiently query the explicit representation for a feature vector.
A tiny MLP receives the feature along with the point coordinates and view direction and regresses an output RGB color for the point.
Figure~\ref{fig:model} shows an overview of the model.

Designing an explicit representation for dynamic 3D scenes is challenging.
Unlike static 3D scenes which are often modeled by point clouds, voxels, or meshes, explicit representations for dynamic scenes have been underexplored.
We show how the key technical challenges of \emph{memory usage} and \emph{sparse observations} can be overcome by our simple HexPlane representation.

\subsection{4D Volumes for Dynamic 3D Scenes}
A dynamic 3D scene could be na\"ively represented as a 4D volume $\mathbf{D}$ comprising independent static 3D volumes per time step $\{\mathbf{V_1}, \mathbf{V_2}, \cdots, \mathbf{V_T}\}$.
However this design suffers from two key problems.
First is \emph{memory consumption}: a na\"ive 4D volume is very memory intensive, requiring $O(N^3 T F)$ space where $N$, $T$, and $F$ are the spatial resolution, temporal resolution, and feature size.
Storing a volume of RGB colors ($F{=}3$) with $N{=}512$, $T{=}32$ in \texttt{float32} format takes 48GB of memory.

The second problem is \emph{sparse observations}.
A single camera moving through a static scene can capture dozens or hundreds of images.
In dynamic scenes capturing multiple images per timestep requires multiple cameras, so we typically have only a few views per timestep;
these sparse views are insufficient for independently modeling each timestep, so we must share information between timesteps.

We reduce \emph{memory consumption} using \emph{factorization}~\cite{chan2022efficient,chen2022tensorf} which has been previously applied to 3D volumes.
We build on TensoRF~\cite{chen2022tensorf} which decomposes a 3D volume $\mathbf{V} \in \mathbb{R}^{XYZF}$\footnote[1]{To simplify notation, we write $\mathbb{R}^{X\times Y}$ as $\mathbb{R}^{XY}$ in this paper.} as a sum of vector-matrix outer products:
\begin{align}
    \begin{split}
        \mathbf{V} =& \sum_{r=1}^{R_1} \mathbf{M}^{XY}_{r} \circ \mathbf{v}^{Z}_{r} \circ \mathbf{v}^{1}_{r} + \sum_{r=1}^{R_2} \mathbf{M}^{XZ}_{r} \circ \mathbf{v}^{Y}_{r} \circ \mathbf{v}^{2}_{r}  \\
        &+ \sum_{r=1}^{R_3} \mathbf{M}^{ZY}_{r} \circ \mathbf{v}^{X}_{r} \circ \mathbf{v}^{3}_{r}
    \end{split}
    \label{eq:vm}
\end{align} 
where $\circ$ is outer product; $\mathbf{M}^{XY}_{r} \circ \mathbf{v}^{Z}_{r} \circ \mathbf{v}^{1}_{r}$ is a low-rank component of $\mathbf{V}$; 
$\mathbf{M}^{XY}_{r} \in \mathbb{R}^{XY}$ is a matrix spanning the $X$ and $Y$ axes,
and $\mathbf{v}^{Z} \in \mathbb{R}^Z, \mathbf{v}^{i}_{r} \in \mathbb{R}^F $ are vectors along the $Z$ and $F$ axes.
$R_1, R_2, R_3$ are the number of low-rank components.
With $R = R_1 + R_2 + R_3\ll N$, this design reduces memory usage from $O(N^3TF)$ to $O(RN^2T)$.

\subsection{Linear Basis for 4D Volume}
Factorization helps reduce memory usage, but factoring an independent 3D volume per timestep still suffers from sparse observations and does not share information across time.
To solve this problem, we can represent the 3D volume $\mathbf{V_t}$ at time $t$ as the weighted sum of a set of shared 3D basis volumes $\{\hat{\mathbf{V}}_{1}, \ldots, \hat{\mathbf{V}}_{R_t}\}$; then
\vspace{-2mm}
\begin{equation}
    \mathbf{V_t} = \sum_{i=1}^{R_t} \mathbf{f}(t)_i \cdot \hat{\mathbf{V}}_{i}
    \label{eq:vb}
\vspace{-2mm}
\end{equation}
where $\cdot$ is a scalar-volume product, $R_t$ is the number of shared volumes,
and $\mathbf{f}(t) \in \mathbb{R}^{R_t}$ gives weights for the shared volumes as a function of $t$.
Shared volumes allow information to be shared across time.
In practice each $\hat{\mathbf{V}}_i$ is represented as a TensoRF as in Equation~\ref{eq:vm} to save memory.

Unfortunately, we found in practice (and will show with experiments) that shared volumes are still too costly; we can only use small values for $R_t$ without exhausting GPU memory.
Since each shared volume is a TensoRF, it has its own independent $\mathbf{M}_r^{XY},\mathbf{v}_r^Z$, \etc;
we can further improve efficiency by sharing these low-rank components across all shared volumes.
The 3D volume $\mathbf{V}_t$ at time $t$ is then

{\footnotesize
\begin{align}
  \mathbf{V}_t =& \sum_{r=1}^{R_1} \mathbf{M}^{XY}_{r} \circ \mathbf{v}^{Z}_{r} \circ \mathbf{v}^{1}_{r} \cdot \mathbf{f}^1(t)_r
  + \sum_{r=1}^{R_2} \mathbf{M}^{XZ}_{r} \circ \mathbf{v}^{Y}_{r} \circ \mathbf{v}^{2}_{r} \cdot \mathbf{f}^2(t)_r \label{eq:vmt} \\[-4mm]
  &+ \sum_{r=1}^{R_3} \mathbf{M}^{ZY}_{r} \circ \mathbf{v}^{X}_{r} \circ \mathbf{v}^{3}_{r} \cdot \mathbf{f}^3(t)_r \nonumber
\end{align}}
where each $\mathbf{f}^i(t)\in\mathbb{R}^{R_i}$ gives a vector of weights for the low-rank components at each time $t$.

In this formulation, $\mathbf{f}^i(t)$ captures the model's dependence on time.
The correct mathematical form for $\mathbf{f}^i(t)$ is not obvious.
We initially framed $\mathbf{f}^i(t)$ as a learned combination of sinusoidal or other fixed basis functions, with the hope that this could make periodic motion easier to learn;
however we found this inflexible and hard to optimize.
$\mathbf{f}^i(t)$ could be an arbitrary nonlinear mapping, represented as an MLP; however this would be slow.
As a pragmatic tradeoff between flexibility and speed, we represent $\mathbf{f}^i(t)$ as a learned piecewise linear function,
implemented by linearly interpolating along the first axis of a learned $T\times R_i$ matrix.

\subsection{HexPlane Representation}
Equation~\ref{eq:vmt} fully decouples the spatial and temporal modeling of the scene: $\mathbf{f}^i(t)$ models time and other terms model space.
However in real scenes space and time are entangled; 
for example a particle moving in a circle is difficult to model under Equation~\ref{eq:vmt} since its $x$ and $y$ positions are best modeled \emph{separately} as functions of $t$.
This motivates us to replace $\mathbf{v}^{Z}_{r} \cdot \mathbf{f}^1(t)_r$ in Equation~\ref{eq:vmt} with a joint function of $t$ and $z$,
similarly represented as a piecewise linear function;
this can be implemented by bilinear interpolation into a learned tensor of shape $Z\times T\times R_1$.
Applying the same transform to all similar terms then gives our HexPlane representation,
which represents a 4D feature volume $V\in\mathbb{R}^{XYZTF}$ as:
\begin{align}
 \mathbf{D} =&
  \sum_{r=1}^{R_1} \mathbf{M}^{XY}_{r} \circ \mathbf{M}_{r}^{ZT} \circ \mathbf{v}^{1}_{r}
   + \sum_{r=1}^{R_2} \mathbf{M}^{XZ}_{r} \circ \mathbf{M}_{r}^{YT} \circ \mathbf{v}^{2}_{r} \label{eq:MM} \\[-4mm]
   &+ \sum_{r=1}^{R_3} \mathbf{M}^{YZ}_{r} \circ \mathbf{M}_{r}^{XT} \circ \mathbf{v}^{3}_{r} \nonumber
\end{align}
where each $\mathbf{M}_r^{AB}\in\mathbb{R}^{AB}$ is a learned plane of features.
This formulation displays a beautiful symmetry, and strikes a balance between representational power and speed.

We can alternatively express a HexPlane as a function $\mathbf{D}$ which maps a point $(x, y, z, t)$ to an $F$-dimensional feature:
{\footnotesize
\begin{align}
  D(x, y, z, t) = (\mathbf{P}^{XYR_1}_{xy\bullet}\odot\mathbf{P}^{ZTR_1}_{zt\bullet})\mathbf{V}^{R_1F} \nonumber \\
  + (\mathbf{P}^{XZR_2}_{xz\bullet}\odot\mathbf{P}^{YTR_2}_{yt\bullet})\mathbf{V}^{R_2F} \label{eq:hexplane-v1} \\
  + (\mathbf{P}^{YZR_3}_{yz\bullet}\odot\mathbf{P}^{XTR_3}_{xt\bullet})\mathbf{V}^{R_3F} \nonumber
\end{align}}%
where $\odot$ is an elementwise product; the superscript of each bold tensor represents its shape, and $\bullet$ in a subscript represents a slice
so each term is a vector-matrix product.
$\mathbf{P}^{XYR_1}$ stacks all $\mathbf{M}^{XY}_r$ to a 3D tensor, and $\mathbf{V}^{R_1F}$ stacks all $\mathbf{v}^1_r$ to a 2D tensor; other terms are defined similarly.
Coordinates $x,y,z,t$ are real-valued, so subscripts denote bilinear interpolation.
This design reduces memory usage to $O(N^2R + NTR + RF)$.

We can stack all $\mathbf{V}^{R_iF}$ into $\mathbf{V}^{RF}$ and rewrite Eq~\ref{eq:hexplane-v1} as
{\footnotesize
\begin{equation}
  [\mathbf{P}^{XYR_1}_{xy\bullet}\odot\mathbf{P}^{ZTR_1}_{zt\bullet}; 
   \mathbf{P}^{XZR_2}_{xz\bullet}\odot\mathbf{P}^{YTR_2}_{yt\bullet};
   \mathbf{P}^{YZR_3}_{yz\bullet}\odot\mathbf{P}^{XTR_3}_{xt\bullet}] \mathbf{V}^{RF}
   \label{eq:hexplane}
\end{equation}}%
where $;$ concatenates vectors.
As shown in Figure~\ref{fig:model}, a HexPlane comprises three pairs of feature planes; each pair has a spatial and a spatio-temporal plane with orthogonal axes (\eg $XY/ZT$).
Querying a HexPlane is fast, requiring just six bilinear interpolations and a vector-matrix product.

\subsection{Optimization}

\begin{figure*}[!thp]
        \centering
        \includegraphics[width=1.0\textwidth]{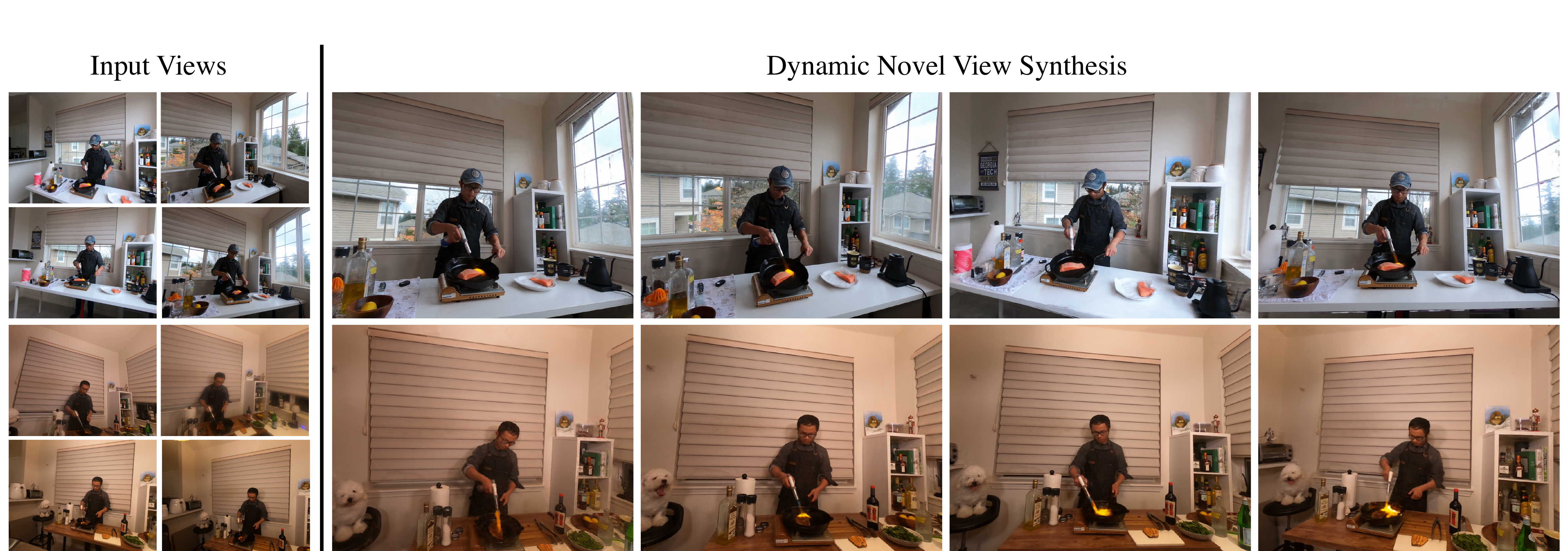}   
        \vspace{-4mm}
        \captionof{figure}{
        \textbf{High-Quality Dynamic Novel View Synthesis on Plenoptic Video dataset~\cite{li2022neural}.}  The proposed HexPlane could effectively represent dynamic 3D scenes with complicated motions and render high-quality results with faithful details at various timesteps and unseen viewpoints.
        We show several samples of input video sequences and synthesis results using a cyclic camera trajectory.}
        \label{fig:syn_results}
        \vspace{-5mm}
        \end{figure*}

We represent dynamic 3D scenes using the proposed HexPlane, which is optimized by photometric loss between rendered and target images. 
For point $(x, y, z, t)$, its opacity and appearance feature are quired from HexPlane, and the final RGB color is regressed from a tiny MLP with appearance feature and view direction as inputs. 
With points' opacities and colors, images are rendered via volumetric rendering. 
The optimization objective is:
\begin{align}
    \mathcal{L} = \frac{1}{| \mathcal{R}|}\sum_{\mathbf{r}\in \mathcal{R}}\|C(\mathbf{r}) - \hat{C}(\mathbf{r})\|^2_{2} + \lambda_{\text{reg}} \mathcal{L}_{\text{reg}}
\end{align}
$\mathcal{L}_{\text{reg}}$, $\lambda_{\text{reg}}$ are regularization and  its weight;
$\mathcal{R}$ is the set of rays and $C(\mathbf{r}), \hat{C}(\mathbf{r})$ are rendered and GT colors of ray $\mathbf{r}$.
\par \noindent {\bf Color Regression.}
To save computations, we query points' opacities directly from one HexPlane, and query appearance features of points with high opacities from another separate HexPlane. 
Queried features and view directions are fed into a tiny MLP for RGB colors. 
An MLP-free design is also feasible with spherical harmonics coefficients as features.

\par \noindent {\bf Regularizer.}
Dynamic 3D reconstruction is a severely ill-posed problem, needing strong regularizers. 
We apply Total Variational~(TV) loss on planes to force the spatial-temporal continuity,
and depth smooth loss in~\cite{niemeyer2022regnerf} to reduce  artifacts.

 \par \noindent {\bf Coarse to Fine Training.}
 A coarse-to-fine scheme is also employed like ~\cite{chen2022tensorf,yu2021plenoctrees}, where the resolution of grids gradually grows during training. 
 This design accelerates the training and provides an implicit regularization on nearby grids. 

\par \noindent {\bf Emptiness Voxel.}
We keep a tiny 3D voxel indicating the emptiness of scene regions and skip points in empty regions. 
Since many regions are empty, it is helpful for acceleration. 
To get this voxel, we evaluate points' opacities across time steps and reduce them to a single voxel with maximum opacities. 
Although keeping several voxels for various time intervals improves speeds, we only keep one for simplicity.

\section{Experiments}
We evaluate HexPlane, our proposed explicit representation, on dynamic novel view synthesis tasks with challenging datasets, comparing its performance and speed to state-of-the-art methods. 
Through extensive ablation studies, we explore its advantages and demonstrate its robustness to different feature fusion mechanisms, coordinate systems, and decoding mechanisms. 
As our objective is to demonstrate the effectiveness of this simple design, we prioritize HexPlane's simplicity and generality without implementing intricate tricks for performance enhancement.

\subsection{Dynamic Novel View Synthesis Results}
For a comprehensive evaluation, we use two datasets with distinct settings: 
the high-resolution, multi-camera \emph{Plenoptic Video dataset}~\cite{li2022neural}, with challenging dynamic content and intricate visual details;
the monocular \emph{D-NeRF dataset}~\cite{pumarola2021d}, featuring synthetic objects. 
\emph{Plenoptic Video dataset} assesses HexPlane's representational capacity for long videos with complex motions and fine details, 
while \emph {D-NeRF dataset}  tests its ability to handle monocular videos and extremely sparse observations (with teleporting~\cite{gao2022dynamic}).

\begin{figure}[t!]
	\centering
    {\includegraphics[width=0.48\textwidth]{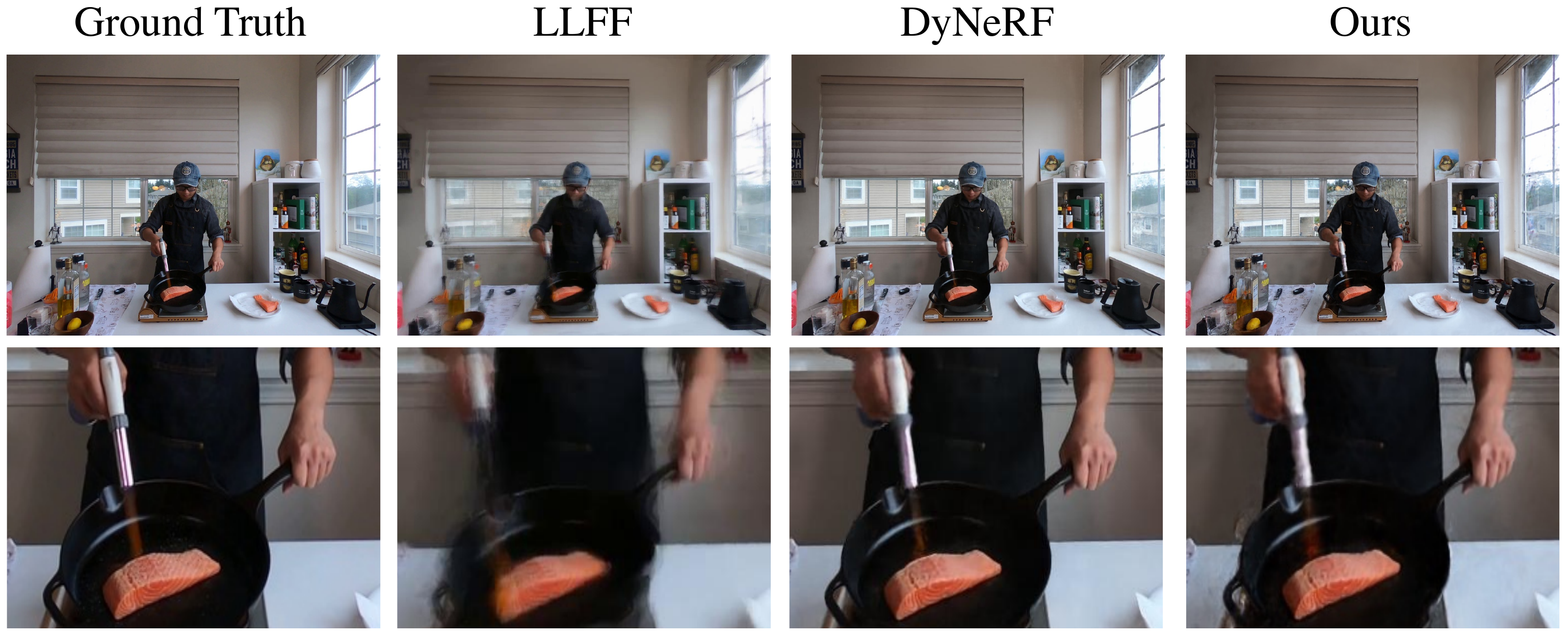}}
	\captionof{figure}{\textbf{Visual Comparison of Synthesis Results.}
	Since DyNeRF~\cite{li2022neural} model is not publicly available, we compare our results to images provided the paper.
	With visually similar results, our proposed HexPlane is over 100 $\times$ faster than DyNeRF. } 
	\label{fig:visual_com}
	\vspace{-6mm}
\end{figure}
\begin{table*}[t!]
    \centering
    \caption{
    \textbf{Quantitative Comparisons on Plenoptic Video dataset~~\cite{li2022neural}.}
    We report synthesis quality, training times (measured in GPU hours) with speedups relative to DyNeRF~\cite{li2022neural}.
    With $672\times$ speedups, HexPlane\dag  with fewer training iterations  has comparable quantitative results to DyNeRF.
    And HexPlane trained with the same iterations noticeably outperforms DyNeRF.
    Baseline methods are evaluated on a particular scene, and we also report average results on all public scenes~(\emph{-all}). 
    \colorbox{lightyellow}{Best} and \colorbox{lightgray}{Second} results are in highlight.}
    \vspace{-1mm}
    \resizebox{0.96\textwidth}{!}{
    \begin{tabular}{p{3cm} P{1.5cm}  P{1.5cm}  P{1.5cm} P{1.5cm} P{1.5cm} P{2.5cm} P{2.0cm}}
    \toprule[2pt]
    Model &Steps & PSNR$\uparrow$  & D-SSIM$\downarrow$ & LPIPS$\downarrow$ & JOD $\uparrow$ &Training Time$\downarrow$ &Speeds-up $\uparrow$\\
    \midrule
    Neural Volumes~\cite{lombardi2019neural} &- & 22.800 & 0.062 &0.295 &6.50 &- &-\\
    LLFF~\cite{mildenhall2019local} & -&23.239  &0.076 &0.235 &6.48 &- &- \\
    NeRF-T~\cite{li2022neural} & - &28.449 &0.023 &0.100 &7.73 & - & - \\
    DyNeRF~\cite{li2022neural}  &650k  & \bestcell{29.581} & \secondcell{0.020} & 0.099 & 8.07 &1344h &1$\times$ \\    
    HexPlane & 650k &\secondcell{29.470} & \bestcell{0.018} &\bestcell{0.078} & \bestcell{8.16} &\secondcell{12h} & \secondcell{$112 \times$}\\
    HexPlane\dag & 100k & 29.263 & \secondcell{0.020} & \secondcell{0.097} &\secondcell{8.14} & \bestcell{2h} & \bestcell{672} $\times$ \\
    \midrule
    HexPlane-all & 650k &31.705 &0.014 & 0.075 & 8.47 & 12h & 112 $\times$\\  
    HexPlane\dag-all & 100k & 31.569 & 0.016 &0.089 & 8.36 & 2h & 672 $\times$ \\
    \bottomrule [2 pt]
    \end{tabular}}
    \label{tab:results_table}
    \vspace{-4mm}
\end{table*}

\begin{table}[t!]
    \centering
    \caption{
    \textbf{Quantitative Results on D-NeRF dataset~\cite{pumarola2021d}.}
    Without deformation, HexPlane has comparable or better results compared to other deformation-based methods,
    and is noticeably faster.}
    \vspace{-2mm}
    \resizebox{0.48\textwidth}{!}{
    \begin{tabular}{  p{2.5cm} P{1cm} P{1.2cm} P{1.2cm} P{1.2cm} P{1.5cm}}
    \toprule[2pt]
    Model & Deform.  &PSNR$\uparrow$  &SSIM $\uparrow$ & LPIPS $\downarrow$ & Training Time$\downarrow$\\
    \midrule
    T-NeRF~\cite{pumarola2021d} & &29.51 & 0.95 & 0.08 &  -\\
    D-NeRF~\cite{pumarola2021d} & \checkmark &30.50 &0.95 & 0.07 & 20 hours\\
    TiNeuVox-S~\cite{fang2022fast} &\checkmark &30.75 & 0.96 & 0.07 & 12m 10s\\
    TiNeuVox-B~\cite{fang2022fast} &\checkmark &\textbf{32.67} & \textbf{0.97} & \textbf{0.04} &49m 46s\\
    HexPlane~(ours) & &31.04 &\textbf{0.97} &\textbf{0.04}&\textbf{11m 30s}\\
    \bottomrule [2 pt]
    \end{tabular}}
    \vspace{-6mm}
    \label{tab:dnerf}
\end{table}

\par \noindent {\bf Plenoptic Video dataset}~\cite{li2022neural}
is a real-world dataset captured by a multi-view camera system using 21 GoPro at 2028 $\times$ 2704 (2.7K) resolution and 30 FPS.
Each scene comprises 19 synchronized, 10-second videos, with 18 designated for training and one for evaluation. 
This dataset is suitable to test the representation ability as it features complex and challenging dynamic content such as highly specular, translucent, and transparent objects; topology changes; moving self-casting shadows; fire flames and strong view-dependent effects for moving objects; and so on.

For a fair comparison, we adhere to the same training and evaluation pipelines as DyNeRF\cite{li2022neural} with slight changes due to GPU resources.   
\cite{li2022neural} trains its model on 8 V100 GPUs for a week, with 24576 batch size for 650K iterations.
We train our model on a single 16GB V100 GPU, with a    4096 batch size and the same iteration numbers, which is $6\times$ fewer sampling.
We follow the same importance sampling design and hierarch training as~\cite{li2022neural}, with 512 spatial grid sizes and 300 time grid sizes.
The scene is in NDC~\cite{mildenhall2020nerf}.

As shown in Figure~\ref{fig:syn_results}, HexPlane delivers high-quality dynamic novel view synthesis across various times and viewpoints. It accurately models real-world scenes with intricate motions and challenging visual features, such as flames, showcasing its robust representational capabilities.

Quantitative comparisons with SOTA methods are in Table~\ref{tab:results_table}, with baseline results from ~\cite{li2022neural} paper.
PSNR, structural dissimilarity index measure (DSSIM)~\cite{sara2019image}, perceptual quality measure LPIPS~\cite{zhang2018unreasonable} and video visual difference measure Just-Objectionable-Difference (JOD)~\cite{mantiuk2021fovvideovdp} are evaluated for comprehensive study. 
Besides results on the ``flame salmon'' scene like ~\cite{li2022neural}, we also report average results on all public scenes except the unsynchronized one, referred to \emph{HexPlane-all}. 
We also train a model with fewer training iterations as \emph{HexPlane\dag}.

As shown in Table~\ref{tab:results_table}, \emph{HexPlane\dag} achieves comparable performance to DyNeRF~\cite{li2022neural} while substantially faster~($672\times$ speedups), highlighting the benefits of employing explicit representations.
DyNeRF uses a giant MLP and per-frame latent codes to represent dynamic scenes, which is slow due to tremendous MLP evaluations.
When trained with the same iteration number, \emph{HexPlane} outperforms DyNeRF in all metrics except PSNR, while being above 100 $\times$ faster.
Although explicit representations typically demand significant memory for their rapid processing speeds due to explicit feature storage, HexPlane occupies a mere 200MB for the entire model. This relatively compact size is suitable for most GPUs. Given its fast speed, we believe this tradeoff presents an attractive option.

Since the model of DyNeRF is not publicly available, it is hard to compare the visual results directly.
We download images from the original paper and find the most matching images in our results,  which are compared in Figure~\ref{fig:visual_com}.

\par \noindent {\bf D-NeRF dataset}~\cite{pumarola2021d} is a monocular video dataset with $360^{\circ}$ observations for synthetic objects.  
Dynamic 3D reconstruction for monocular video is challenging since only one observation is available each time. 
Current SOTA methods for monocular video usually have a deformation field and a static canonical field, where points in dynamic scenes are mapped to positions in the canonical field.
The mapping~(deformation) is represented by another MLP.

The underlying assumption of deformation field design is that there are no topology changes, which does not always hold in the real world while holding in this dataset.
Again, to keep HexPlane general enough, we do not assume deformation, the same as T-NeRF~\cite{pumarola2021d}.
We use this dataset to validate the ability to work with monocular videos.

We show quantitative results in Table~\ref{tab:dnerf}.
For fairness, all training times are re-measured on the same 2080TI GPU.
Our HexPlane distinctly outperforms other methods even without introducing the deformation field,  demonstrating the inherent ability to deal with sparse observations due to the shared basis.
Again, our method is hundreds of times faster than MLP-based designs like D-NeRF and T-NeRF. 

Tineuvox~\cite{fang2022fast} is a recent work for accelerating D-NeRF, replacing canonical space MLP with a highly-optimized sparse voxel Cuda kernel and keeping an MLP to represent deformation.
Therefore, it still uses explicit representation for static scenes while our target is dynamic scenes. 
Without any custom Cuda kernels, our method is faster and better than its light version and achieves the same LPIPS and SSIM as its bigger version, which takes longer time to train. 

\subsection{Ablations and Analysis}
We run deep introspections to HexPlane by answering questions with extensive ablations.
Ablations are conducted mainly on D-NeRF~\cite{pumarola2021d} dataset because of efficiency.
\begin{table}[t!]
    \centering
    \caption{
    \textbf{Quantitative Results for Different Factorizations.}
    Various factorization designs are evaluated on D-NeRF dataset with different $R$~(basis number).
    HexPlane achieves the best quality and speed among all methods.}
    \vspace{-2mm}
    \resizebox{0.48\textwidth}{!}{
    \begin{tabular}{p{2.5cm} P{1.2cm} P{1.0cm} P{1.0cm} P{1.0cm} P{1.25cm}}
    \toprule[2pt]
    Model &\centering $R$ &PSNR$\uparrow$  & SSIM$\uparrow$ & LPIPS$\downarrow$ &Training Time$\downarrow$ \\
    \midrule
    \multirow{3}{*}{Volume Basis} &8  & 30.460 & 0.965 & 0.045 &\textbf{18m 04s} \\
    &12 &30.587 &0.966 &0.043  &24m 06s \\
    &16  & \textbf{30.631} & \textbf{0.967} & \textbf{0.042} &29m 20s\\
    \midrule
    \multirow{3}{*}{VM-T} &24 &30.329  &0.962 &0.051 & \textbf{14m 36s} \\
    &48 &30.657 &0.965 & 0.048 &15m 58s\\
    &96 & \textbf{30.744} & \textbf{0.966} & \textbf{0.045} &17m 03s\\
    \midrule
    \multirow{4}{*}{CP Decom.} &48 &28.370 &0.942 &0.083 & \textbf{10m 31s}\\
    &96 &29.371 &0.951 & 0.070 & 11m 03s \\
    &192 &30.086 & 0.957 & 0.063 & 11m 33s \\
    &384 & \textbf{30.302} & \textbf{0.959} & \textbf{0.059} & 13m 06s\\
    \midrule   
    \multirow{2}{*}{HexPlane} &24 &30.886 &0.966 &0.042 & \textbf{10m 27s} \\
    &48 &\textbf{31.042} & \textbf{0.968} & \textbf{0.039} & 11m 30s\\
    \bottomrule [2 pt]
    \end{tabular}}
    \vspace{-3mm}
    \label{tab:factoring_ablation}
\end{table}

\par \noindent {\bf How does HexPlane compare to others?}
We compare HexPlane with other designs mentioned in the \emph{Method Section}  in Table~\ref{tab:factoring_ablation}, where each method has various basis numbers $R$:
(1). \emph{Volume Basis} represents 4D volumes as weighted summation of a set of shared 3D volumes as Eq~\ref{eq:vb}, which 3D volume is represented as Eq~\ref{eq:vm};
(2). \emph{VM-T}~(vector, matrix and time) uses Eq~\ref{eq:vmt} representing 4D volumes;
(3). \emph{CP Decom.}~(CANDECOMP Decomposition) follows ~\cite{chen2022tensorf}, which represents 4D volumes using a set of vectors for each axis.
Implementation details are shown in Supp.

HexPlane gives optimal performance among all methods, illustrating the advantages of spatial-temporal planes. 
Compared to other methods, spatial-temporal planes allow HexPlane to model motions effectively with a small basis number $R$, leading to higher efficiency as well.
Increasing $R$ used for representation leads to better results while also resulting in more computations. 
We also notice that an unsuitable large $R$ may lead to the overfitting problem, which instead harms synthesis quality on novel views. 

\par \noindent {\bf Could variants of HexPlane work?}
HexPlane has excellent symmetry as it contains all pairs of coordinate axes. 
By breaking this symmetry, we evaluate other variants in Table~\ref{tab:six_plane_ablation}.
\emph{Spatial Planes} only have three spatial planes: $\mathbf{P}^{XY}, \mathbf{P}^{XZ}, \mathbf{P}^{YZ}$, and
\emph{Spatial-Temporal Planes} contain the left three spatial-temporal planes;
\emph{DoublePlane} contains only one group of paired planes, i.e. $\mathbf{P}^{XY}, \mathbf{P}^{ZT}$;
\emph{HexPlane-Swap} groups planes with repeated axes like $\mathbf{P}^{XY}, \mathbf{P}^{XT}$.
We report their performance and speeds. 
\begin{table}[t!]
    \centering
    \caption{
    \textbf{Ablations on Feature Planes Designs.}
    We remove and swap HexPlane's planes and show results on D-NeRF dataset. }
    \vspace{-2mm}
    \resizebox{0.48\textwidth}{!}{
    \begin{tabular}{p{3.5cm} P{1.3cm} P{1.3cm} P{1.3cm} P{1.3cm} }
    \toprule[2pt]
    Model  &PSNR$\uparrow$  & SSIM$\uparrow$ & LPIPS$\downarrow$ &Training Time$\downarrow$\\
    \midrule
    Spatial Planes &20.369 &0.879 & 0.148 &\secondcell{9m 02s} \\
    Spatial-Temporal Planes & 21.112 & 0.879 & 0.148 &9m 29s\\
    DoublePlane~(XY-ZT) & \secondcell{30.370} & \secondcell{0.961} & \secondcell{0.054} & \bestcell{8m 04s}\\ 
    HexPlane-Swap & 28.562 &0.954 & 0.056 & 11m 44s\\  
    HexPlane &\bestcell{31.042} & \bestcell{0.968} & \bestcell{0.039} & 11m 30s\\
    \bottomrule [2 pt]
    \end{tabular}}
    \vspace{-5mm}
    \label{tab:six_plane_ablation}
\end{table}

\begin{table}[thp!]
    \centering
    \caption{
    \textbf{Ablations on Feature Fusions Designs.}
    We show results with various fusion designs on D-NeRF dataset.
    HexPlane could work with other fusion mechanisms, showing its robustness.}
    \vspace{-3mm}
    \resizebox{0.48\textwidth}{!}{
    \begin{tabular}{p{1.5cm} p{1.5cm} P{1.5cm} P{1.5cm} P{1.5cm} P{1.5cm}}
    \toprule[2pt]
    Fusion-One &Fusion-Two &PSNR$\uparrow$  & SSIM$\uparrow$ & LPIPS$\downarrow$ \\
    \midrule
    \multirow{3}{*}{Multiply} & Concat & \bestcell{31.042}   &\bestcell{0.968} & \bestcell{0.039} \\
    & Sum & \secondcell{31.023} & \secondcell{0.967} & \bestcell{0.039} \\
    & Multiply & 30.345  & 0.966 & 0.041 \\
    \midrule
    \multirow{3}{*}{Sum} & Concat  & 25.428 & 0.931 &0.084   \\
    & Sum &25.227 &0.928 &0.090  \\
    & Multiply& 30.585 & 0.965 &0.044  \\
    \bottomrule [2 pt]
    \end{tabular}}
    \vspace{-4mm}
    \label{tab:fusion_abalation}
\end{table}

\begin{figure}[thp!]
        \centering
        \includegraphics[width=0.48\textwidth]{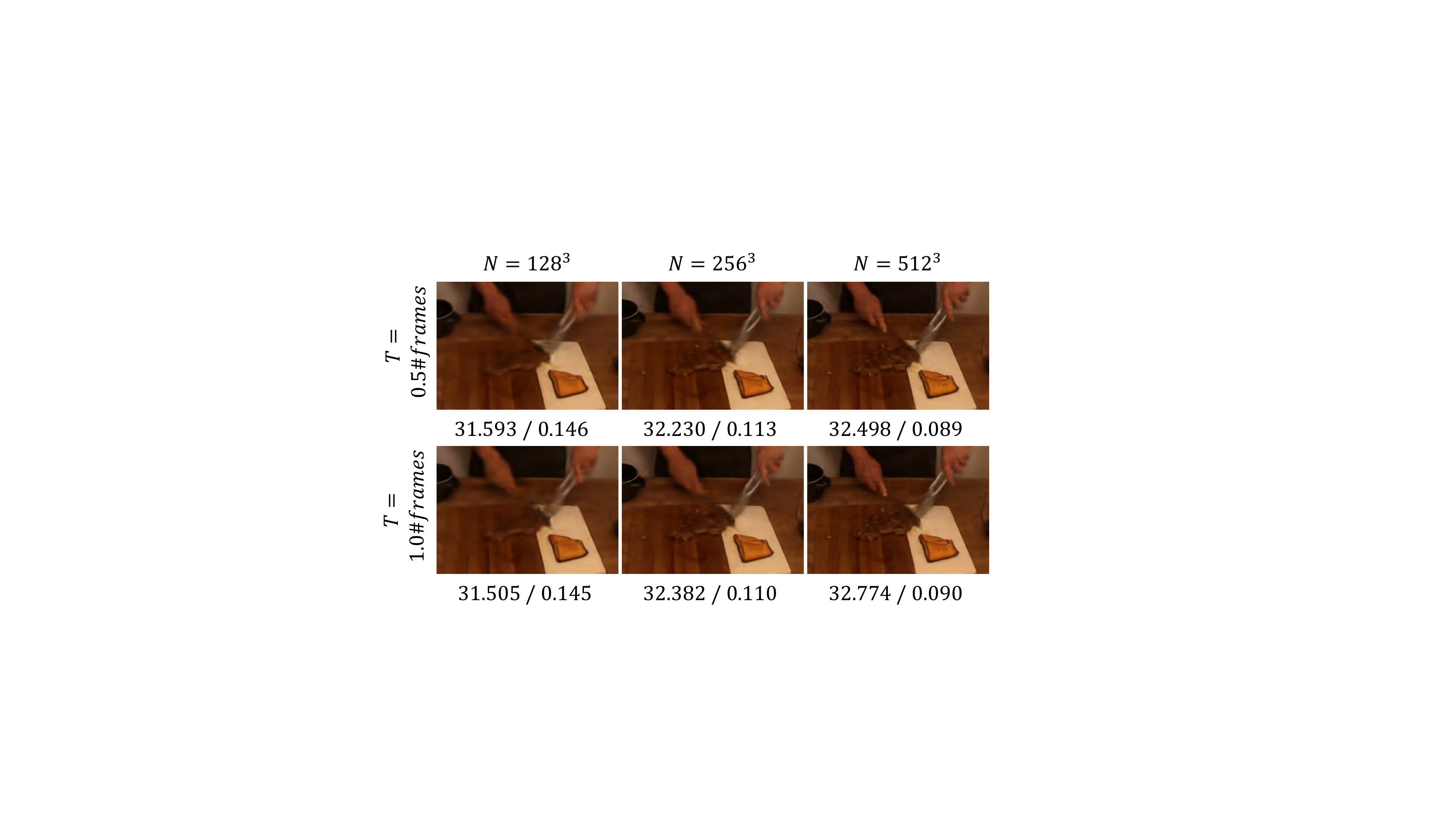}
        \vspace{-7mm}
        \caption{
        \textbf{Synthesis Results with Different Spacetime Grid Resolutions.}
        We show zoomed in synthesis results on Plenoptic Video dataset with space grid resolution ranging from $128^3$ to $512^3$ and 
        time grid ranging from half to one of the video frame number.
        PSNR and LPIPS of the scene are reported below each images.
        }
        \label{fig:grid_aba}
        \vspace{-6mm}
    \end{figure}

As shown in the table, neither \emph{Spatial Planes} nor \emph{Spatial-Temporal Planes} could represent dynamic scenes alone, indicating both are essential for representations.
\emph{HexPlane-Swap} achieves inferior results since its axes are not complementary, losing features from the particular axis.  
\emph{DoublePlane} is less effective than HexPlane since HexPlane contains more comprehensive spatial-temporal modes. 

\par \noindent {\bf How does grid resolution affect results?}
We show qualitative results with various spacetime grid resolutions in Figure~\ref{fig:grid_aba} and report its PSNR/LPIPS below zoomed-in images.
Besides the space grid ranging from $128^3$ to $512^3$, we compare results with different time grid resolutions, ranging from half to the same as video frames.
Higher resolutions of the space grid lead to better synthesis quality, shown by both images and metrics.  
HexPlane results are not noticeably affected by a smaller time grid resolution.

\subsection{Robustness of HexPlane Designs}
In addition to its performance and efficiency, this section demonstrates HexPlane's robustness to diverse design choices, resulting in a highly adaptable and versatile framework. This flexibility allows for its applications across a wide range of tasks and research directions.
\par \noindent {\bf Various Feature Fusion Mechanisms.}
In HexPlane, feature vectors from each plane are extracted and subsequently fused into a single vector, which are multiplied by matrix $\mathbf{V}^{RF}$ later for final results.
During fusion, features from paired planes are first element-wise multiplied~(\emph{fusion one}) and then concatenated into a single one~(\emph{fusion two}). 
We explore other fusion designs beyond this \emph{Multiply-Concat}.
\begin{table}[t!]
    \centering
    \caption{
    \textbf{Dynamic View Synthesis without MLPs.}
    HexPlane-SH is a pure explicit model without MLPs on D-NeRF dataset, which stores spherical harmonics~(SH) as appearance features and directly regress RGB from it rather than MLPs.
    HexPlane-SH gives reasonable results and faster than HexPlane with MLP.
    }
    \vspace{-2mm}
    \resizebox{0.48\textwidth}{!}{
    \begin{tabular}{p{2.5cm} P{1.3cm} P{1.3cm} P{1.3cm} P{3cm} }
    \toprule[2pt]
    Model  &PSNR$\uparrow$  & SSIM$\uparrow$ & LPIPS$\downarrow$ &Training Time$\downarrow$\\
    \midrule 
    HexPlane &31.042 &0.968 & 0.039 & 11m 30s\\
    HexPlane-SH & 29.284 & 0.952 & 0.056 & 10m 42s\\
    \bottomrule [2 pt]
    \end{tabular}}
    \vspace{-3mm}
    \label{tab:SH_aba}
\end{table}

\begin{figure}[!thp]
    \centering
    \resizebox{0.48\textwidth}{!}{
    \begin{tabular}{@{}c@{\hspace{0.5mm}}c@{\hspace{0.5mm}}c@{\hspace{0.5mm}}c@{}}
    \includegraphics[width=0.12\textwidth]{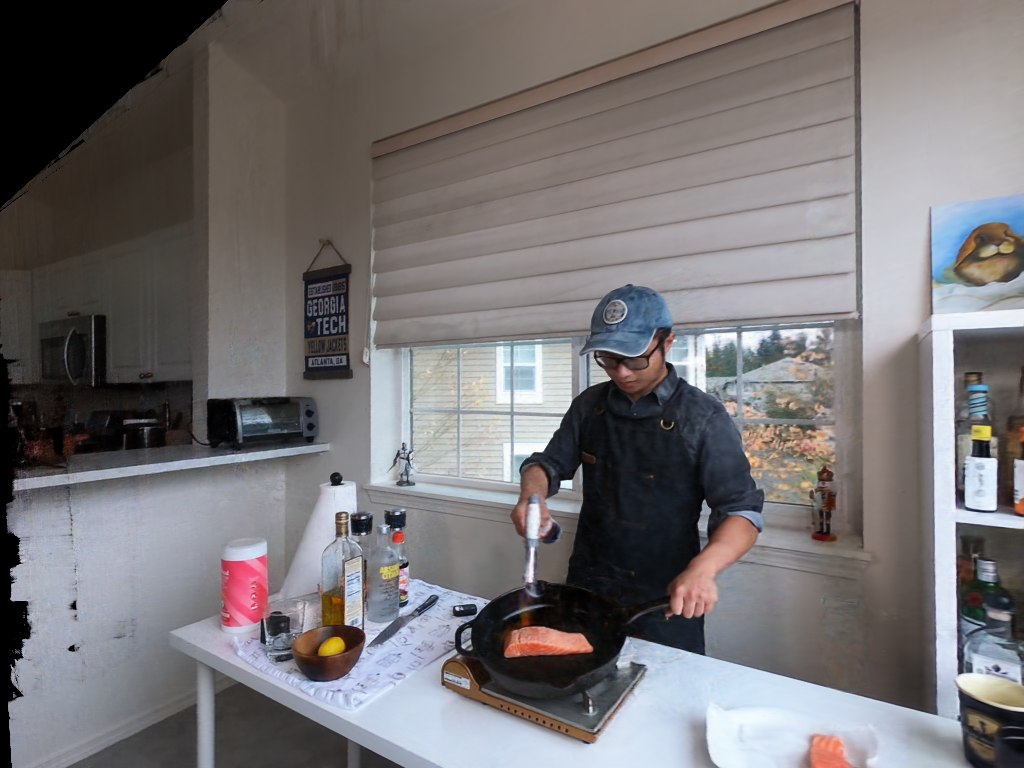} &
    \includegraphics[width=0.12\textwidth]{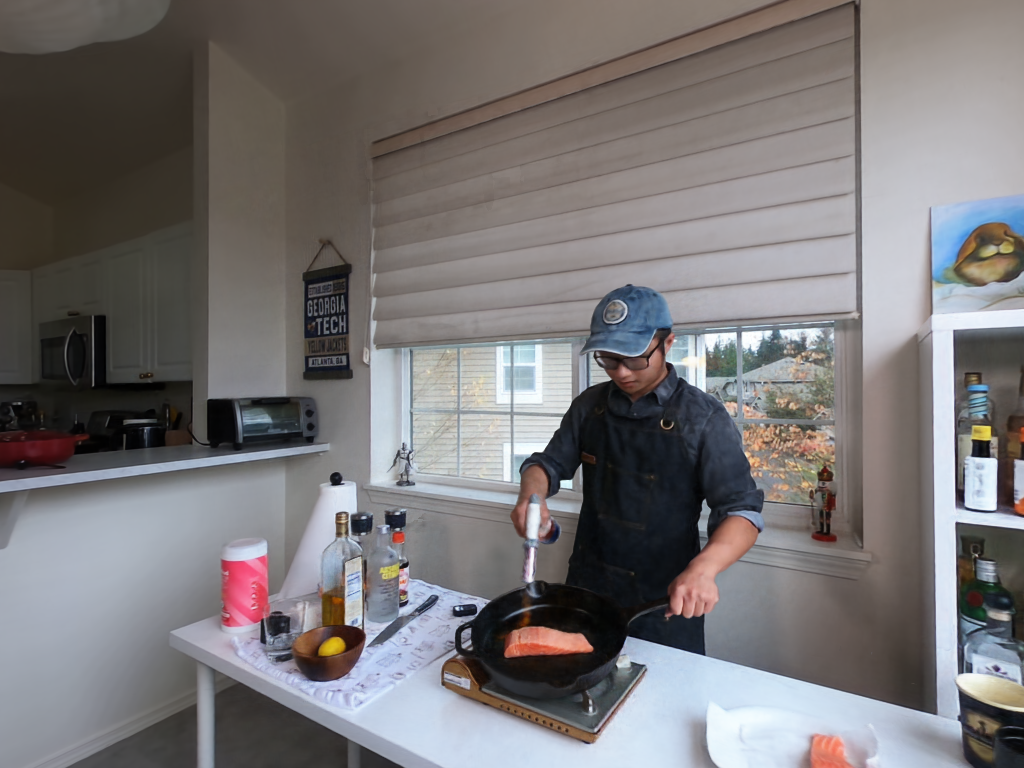} &
    \includegraphics[width=0.12\textwidth]{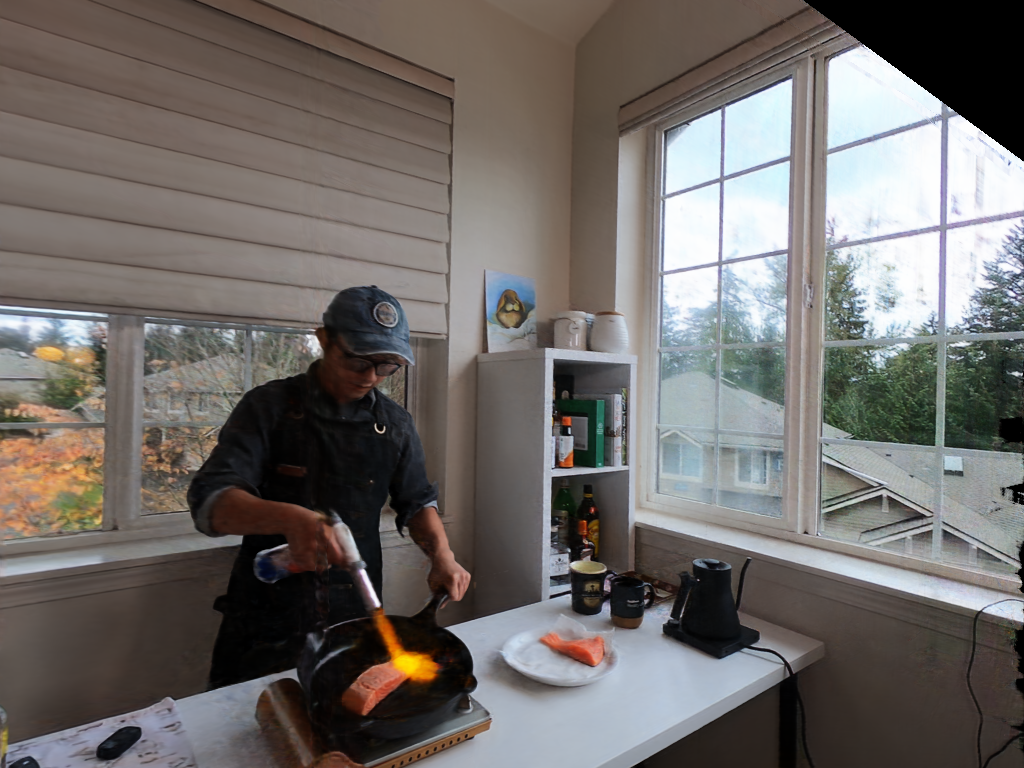} &
    \includegraphics[width=0.12\textwidth]{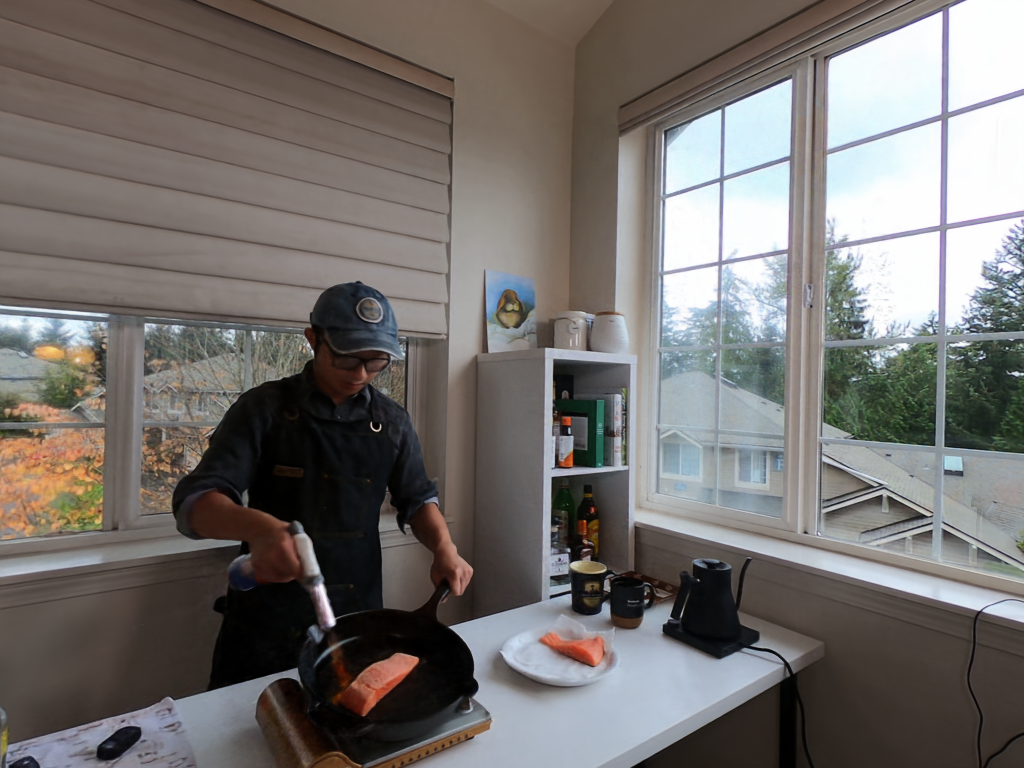}\\
    \small{NDC} & \small{Spherical} & \small{NDC} & \small{Spherical} \\
    \end{tabular}}
    \vspace{-2mm}
    \captionof{figure}{
    \textbf{Synthesis Results at Extreme Views for NDC and Spherical Coordinates.}
    Scenes represented in NDC are assumed to be bounded along $x,y$ axes, whose boundaries are observable at extreme views(top-left and top-right corners), leading to incorrect geometries and artifacts.
    Using spherical coordinate, our HexPlane could seamlessly represent dynamic unbounded scenes.} 
    \label{fig:spherical}
    \vspace{-6mm}
    \end{figure}

Table~\ref{tab:fusion_abalation} shows that \emph{Multiply-Concat} is not the sole viable design.
\emph{Sum-Multiply} and swapped counterpart \emph{Multiply-Sum} both yield good results, albeit not optimal,\footnote{Further tuning of initialization/other factors may lead to better results.}
 highlighting an intriguing symmetry between multiplication and addition.
\emph{Multiply-Multiply} also produces satisfactory outcomes, while \emph{Sum-Sum} or \emph{Sum-Concat} fail, illustrating the capacity limitations of addition compared to multiplication. 
Overall, HexPlane is remarkably robust to various fusion designs.
We show complete results and analysis in Supp.

\par \noindent {\bf Spherical Harmonics Color Decoding.}
Instead of regressing colors from MLPs, we evaluate a pure explicit model in Table~\ref{tab:SH_aba} without MLPs. 
Spherical harmonics~(SH)~\cite{yu2021plenoctrees} coefficients are computed directly from HexPlanes, and decoded to RGBs with view directions.
Using SH allows faster rendering speeds at a slightly reduced quality.
We find that optimizing SH for dynamic scenes is more challenging compared to ~\cite{fridovich2022plenoxels,chen2022tensorf}, which is an interesting future direction.

\begin{figure}[thp!]
    \centering
    \begin{tabular}{@{}c@{\hspace{0.25mm}}c@{\hspace{0.25mm}}c@{\hspace{0.25mm}}c@{}}
    \includegraphics[width=0.12\textwidth]{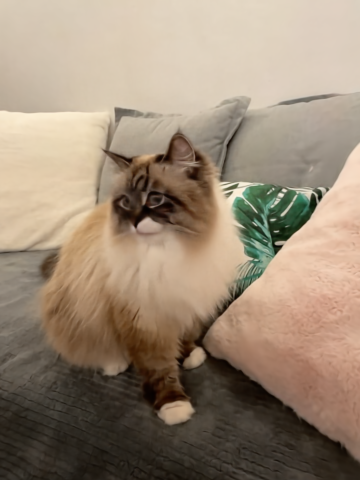} &
    \includegraphics[width=0.12\textwidth]{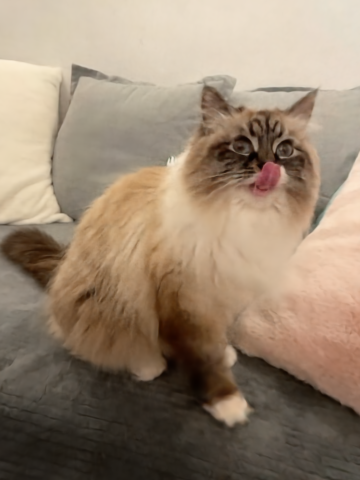} &
    \includegraphics[width=0.12\textwidth]{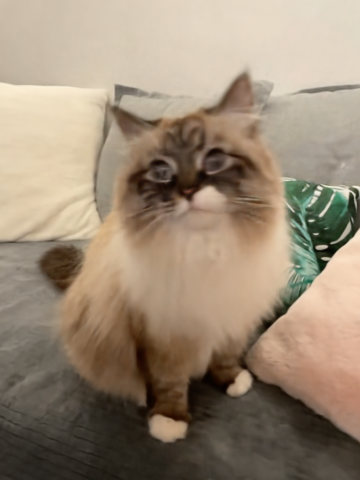} &
    \includegraphics[width=0.12\textwidth]{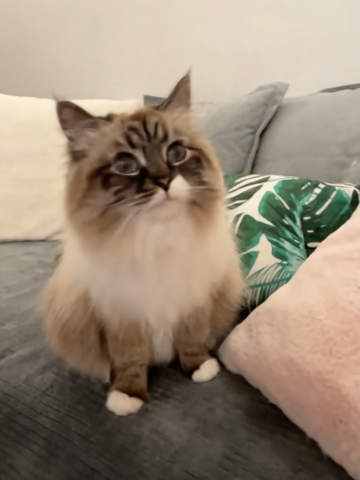} \\
    \includegraphics[width=0.12\textwidth]{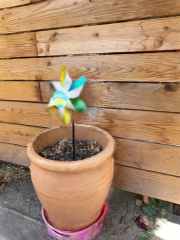} &
    \includegraphics[width=0.12\textwidth]{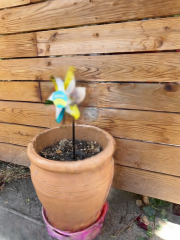} &
    \includegraphics[width=0.12\textwidth]{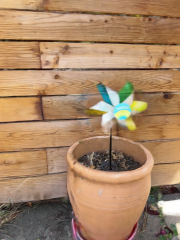} &
    \includegraphics[width=0.12\textwidth]{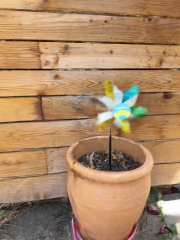} \\
    \end{tabular}
    \vspace{-3mm}
    \caption{
    \textbf{Dynamic Novel View Synthesis on Videos Captured by iPhone.
    }
    We test HexPlane on casual videos captured by iPhone~\cite{gao2022dynamic} and show synthesis results across novel timesteps and views.
    \emph{Row one} are results with interpolated camera poses, while \emph{Row two} shows results with extrapolated viewpoints, which are significantly distinct from camera poses used for video captures.
    }
    \vspace{-5mm}
    \label{fig:iphone}
\end{figure}

\par \noindent {\bf Spherical Coordinate for Unbounded Scenes.}
HexPlane is limited to bounded scenes because grid sampling fails for out-of-boundary points, which is a common issue among explicit representations. 
Even normalized device coordinates~(NDC)~\cite{mildenhall2020nerf} still require bounded $x, y$ values and face-forwarding assumptions. 
This limitation constrains the usage for real-world videos, leading to artifacts and incorrect geometries as shown in Figure~\ref{fig:spherical}.

To address it, we re-parameterize $(x,y,z,t)$ into spherical coordinate $(\theta, \phi, r,t )$ and build HexPlane with $\theta, \rho, r, t$ axes,
where $r=1/\sqrt{x^2 + y^2 + z^2}$, $\theta, \phi$ is the polar angle and azimuthal angle.
During rendering, points are sampled with $r$ linearly placed between 0 and 1. 
Without any special adjustments, HexPlane can represent dynamic fields with spherical coordinates, and deliver satisfactory results,
which provides a solution for modeling unbounded scenes and exhibits robustness to different coordinate systems.

\subsection{View Synthesis Results on Real Captured Video}

We test HexPlane with monocular videos captured by iPhone from ~\cite{gao2022dynamic}, whose camera trajectories are relatively casual and closer to real-world use cases.
We show synthesis results in ~\ref{fig:iphone}.
Without any deformation or category-specific priors, our method could give realistic synthesis results on these real-world monocular videos, faithfully modeling static backgrounds, casual motions of cats, typology changes (cat's tongue), and fine details like cat hairs. 

\section{Conclusion}
We propose HexPlane, an explicit representation for dynamic 3D scenes using six feature planes, which computes features of spacetime points via sampling and fusions.
Compared to implicit representations, it could achieve comparable or even better synthesis quality for dynamic novel view synthesis, with over hundreds of times accelerations.

In this paper, we aim to keep HexPlane neat and general, preventing introducing deformation, category-specific priors, or other specific tricks.
Using these ideas to make HexPlane better and faster would be an appealing future direction.
Also, using HexPlane in other tasks except for dynamic novel view synthesis, e.g., spatial-temporal generation, would be interesting to explore~\cite{Singer2023TextTo4DDS}. 
We hope HexPlane could contribute to a broad range of research in 3D.

\vspace{1mm}\noindent\textbf{Acknowledgments}
Toyota Research Institute provided funds to support this work but this article solely reflects the opinions and conclusions of its authors and not TRI or any other Toyota entity.
We thank Shengyi Qian for the title suggestion, David Fouhey, Mohamed El Banani, Ziyang Chen, Linyi Jin and  for helpful discussions and feedbacks.
\clearpage
{\small
\bibliographystyle{ieee_fullname}
\bibliography{egbib}
}

\section{General Discussions}
\subsection{Broader Impacts}
Our work aims to design an explicit representation for dynamic 3D scenes.
We only reconstruct existing scenes and render images from different viewpoints and timesteps.
Therefore, we don't generate any new scenes or deceive contents which don't exist before.
Our current method is not intended and can also not be used to create fake materials, which could mislead others. 

We use \emph{Plenoptic Video dataset}~\cite{li2022neural} in our experiments, which contains human faces in videos.
This dataset is a public dataset with License: CC-BY-NC 4.0 with consent.

Our method is hundreds of times faster than existing methods, consuming significantly less computation resources. 
Considering the GPU resource usages, our method could save considerably carbon emission.

\subsection{Limitations and Future Directions}
For comprehensively understanding HexPlane, we discuss its limitations and potential future improvements. 

The ultimate goal of our paper is to propose and validate an explicit representation for dynamic scenes instead of purchasing SOTA numbers. 
To this end, we intend to make HexPlanes simple and general, making minor assumptions about the scenes and not introducing complicated tricks to improve performance. 
This principle leads to elegant solutions while potentially limiting performance as well. 
In the following, we will discuss these in detail. 

Many methods~\cite{park2021hypernerf,park2021nerfies,pumarola2021d,fang2022fast} use deformation and canonical fields to represent dynamic 3D scenes with monocular videos, where spacetime points are mapped into a static 3D scene represented by a canonical field.  
Again, we don't employ deformation fields in our design since this assumption is not always held in the real world, especially for scenes with typology changes and new-emerging content. 
But this design is very effective for monocular videos since it introduces a solid information-sharing mechanism and allows learning 3D structures from very sparse views. 
HexPlane uses an inherent basis-sharing mechanism to cope with sparse observations.
Although this design is shown to be powerful in our experiments, it is still less effective than the aforementioned deformation field, leading to degraded results for scenes with extremely sparse observations. 
Introducing deformation fields into HexPlane, like using HexPlane to represent deformation fields, would be an appealing improvement for monocular videos.

Similarly, category-specific priors like 3DMM~\cite{egger20203d} or SMPL~\cite{SMPL:2015} are even more powerful than deformation fields, which enormously improve results but are hardly limited to particular scenes. 
Combining these ideas with HexPlane for specific scenes would be very interesting.

\begin{figure*}[!thp]
    \centering
    \includegraphics[width=0.95\textwidth]{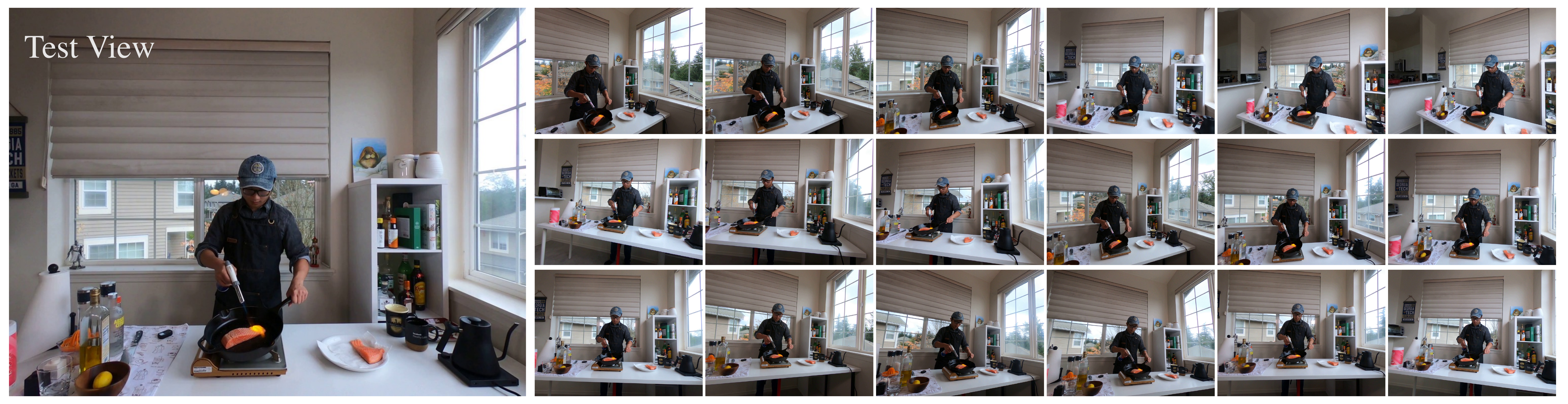}
	\captionof{figure}{{\bf Train and Test View of Plenoptic Video Dataset~\cite{li2022neural}.}
    Plenoptic Video Dataset has 18 train views and 1 test view.} 
	\label{fig:nv3d_view}
\end{figure*}
Existing works demonstrated that explicit representations are prone to giving artifacts and require strong regularizations for good results, which also holds in HexPlane. 
There are color jittering and artifacts in the synthesized results, demanding stronger regularizations and other tricks to improve results further. 
Special spacetime regularizations and other losses like optical flow loss would be an interesting future direction to explore. 
Also, instead of simply representing everything using spherical coordinates in the paper, we could have a foreground and background model like NeRF++~\cite{zhang2020nerf++}, where the background is modeled in spherical coordinates. 
Having separate foreground background models could noticeably improve the results. 
Moreover, rather than using the same basis for representing a long video, using a different basis for different video clips may give better results. 
We believe HexPlane could be further improved with these adjustments.

Besides dynamic novel view synthesis, we believe HexPlane could be utilized in a broader range of research, like dynamic scene generation or edits. 
\subsection{License}

We provide licenses of assets used in our paper.

\par \noindent \textbf{Plenoptic Video Dataset~\cite{li2022neural}.}
We evaluate our method on all all public scenes of Plenoptic Video dataset~\cite{li2022neural}, except a-synchronize scene ``coffee-martini''.
The dataset is in \url{https://github.com/facebookresearch/Neural_3D_Video} with License  CC-BY-NC 4.0 

\par \noindent \textbf{D-NeRF Dataset~\cite{pumarola2021d}.}
We use D-NeRF dataset provided in \url{https://github.com/albertpumarola/D-NeRF}.

\par \noindent \textbf{iPhone Dataset~\cite{gao2022dynamic}.}
The iPhone dataset is provided in \url{https://github.com/KAIR-BAIR/dycheck/}, licensed under  Apache-2.0 license.

\par \noindent \textbf{Plenoptic Video Dataset Baselines~\cite{li2022neural}.}
For all baselines in this dataset, we use numbers reported in the original paper since these models are not publicly available.

\par \noindent \textbf{D-NeRF Dataset Baselines~\cite{pumarola2021d}.}
D-NeRF model is in \url{https://github.com/albertpumarola/D-NeRF} and 
Tineuvox model~\cite{fang2022fast} is in \url{https://github.com/hustvl/TiNeuVox}.
Tineuvox is licensed under Apache License 2.0.

\section{Training Details and More Results}

\subsection{Plenoptic Video Dataset~\cite{li2022neural}.}
Plenoptic Video Dataset~\cite{li2022neural} is a multi-view real-world video dataset, where each video is 10-second long.
The training and testing views are shown in Figure~\ref{fig:nv3d_view}.

We have $R_1 = 48, R_2 = 24, R_3 = 24$ for appearance HexPlance, where $R_1, R_2, R_3$ are basis numbers for $XY-ZT, XZ-YT, YZ-XT$ planes.
For opacity HexPlane, we set $R_1 = 24, R_2 = 12, R_3 = 12$. 
We have different $R_1, R_2, R_3$ since scenes in this dataset are almost face-forwarding, demanding better representation along the $XY$ plane.
The scene is modeled using \emph{normalized device coordinate}~(NDC)~\cite{mildenhall2020nerf} with min boundaries $[-2.5, -2.0, 0.0]$ and max boundaries $[2.5, 2.0, 1.0]$.

Instead of giving the same grid resolutions along $X, Y, Z$ axes, we adjust them based on their boundary distances.
That is, we give larger grid resolution to axis ranging a longer distance, like $X$ axis from $-2.5$ to $2.5$,
and provide smaller grid resolution to axis going a shorter length, like $Z$ axis from $0$ to $1$.
The ratio of grid resolutions for different axes is the same as their distance ratios, while the total grid size number is manually controlled. 

During training, HexPlane starts with a space grid size of $64^3$ and doubles its resolution at 70k, 140k, and 210k to $512^3$.
The emptiness voxel is calculated at 50k and 100k iterations. 
The learning rate for feature planes is 0.02, and the learning rate for $\mathbf{V}^{RF}$ and neural network is 0.001.
All learning rates are exponentially decayed.  
We use Adam~\cite{kingma2014adam} for optimization with $\beta_1 = 0.9, \beta_2 = 0.99$.
We apply Total Variational loss on all feature planes with $\lambda = 0.0005$ for spatial axes and $\lambda=0.001$ for temporal axes.  
\begin{table*}[!thp]
    \centering
    \caption{
    \textbf{Results of Plenoptic Video Dataset~\cite{li2022neural}.}
    We report results of each scene. }
    \resizebox{0.95\textwidth}{!}{
    \begin{tabular}{  p{2.0cm}| P{1.2cm} P{1.5cm} P{1.2cm} P{1.2cm} | P{1.2cm} P{1.5cm} P{1.2cm} P{1.2cm} |P{1.2cm} P{1.5cm} P{1.2cm} P{1.2cm}}
    \toprule[2pt]
    Model &\multicolumn{4}{c}{Flame Salmon} &\multicolumn{4}{c}{Cook Spinach} &\multicolumn{4}{c}{Cut Roasted Beef}  \\
     &PSNR$\uparrow$  &D-SSIM $\downarrow$ & LPIPS $\downarrow$ &  JOD $\uparrow$ &PSNR$\uparrow$  &D-SSIM $\downarrow$ & LPIPS $\downarrow$ &  JOD $\uparrow$ &PSNR$\uparrow$  &D-SSIM $\downarrow$ & LPIPS $\downarrow$ &  JOD $\uparrow$\\
     \midrule
     HexPlane & 29.470 & 0.018 &0.078 &8.16 &32.042 &0.015 & 0.082 & 8.32 & 32.545&0.013 & 0.080 & 8.59\\
    HexPlane\dag &29.263 & 0.020 & 0.097 &8.14 &31.860 & 0.017 & 0.097 & 8.25 & 32.712 & 0.015 & 0.094 & 8.37\\
    \bottomrule [2 pt]
    Model &\multicolumn{4}{c}{Flame Steak} &\multicolumn{4}{c}{Sear Steak} &\multicolumn{4}{c}{Average}  \\
    &PSNR$\uparrow$  &D-SSIM $\downarrow$ & LPIPS $\downarrow$ &  JOD $\uparrow$ &PSNR$\uparrow$  &D-SSIM $\downarrow$ & LPIPS $\downarrow$ &  JOD $\uparrow$ &PSNR$\uparrow$  &D-SSIM $\downarrow$ & LPIPS $\downarrow$ &  JOD $\uparrow$\\
    \midrule
    HexPlane &32.080 & 0.011 & 0.066 & 8.61 & 32.387 & 0.011 & 0.070 & 8.66 & 31.705 & 0.014 & 0.075 & 8.47\\
    HexPlane\dag &31.924 & 0.012 & 0.081 &8.51 &32.085 & 0.014 & 0.079 & 8.51 & 31.569 & 0.016 & 0.090 & 8.36\\
    \bottomrule [2 pt]
    \end{tabular}}
    \label{sup:tab_nv3d}
\end{table*}

\begin{table*}[!thp]
    \centering
    \caption{
    \textbf{Per-Scene Results of D-NeRF Dataset~\cite{pumarola2021d}.}
    We report results of each scene. }
    \resizebox{0.95\textwidth}{!}{
    \begin{tabular}{  p{2.0cm}| P{1.2cm} P{1.2cm} P{1.2cm} | P{1.2cm} P{1.2cm} P{1.2cm} |P{1.2cm}P{1.2cm} P{1.2cm}}
    \toprule[2pt]
    Model &\multicolumn{3}{c}{Hell Warrior} &\multicolumn{3}{c}{Mutant} &\multicolumn{3}{c}{Hook}  \\
    & PSNR$\uparrow$ &SSIM$\uparrow$ &LPIPS $\uparrow$    & PSNR$\uparrow$ &SSIM$\uparrow$ &LPIPS $\uparrow$    & PSNR$\uparrow$ &SSIM$\uparrow$ &LPIPS $\uparrow$\\
    \midrule
    T-NeRF &23.19 & 0.93 & 0.08 & 30.56 & 0.96 & 0.04 & 27.21 & 0.94  & 0.06 \\
    D-NeRF &25.02 &0.95 & 0.06 &31.29 & 0.97 & 0.02 &29.25 & 0.96 & 0.11 \\
    TiNeuVox-S &27.00 & 0.95 & 0.09 & 31.09 & 0.96 & 0.05 & 29.30 &0.95 & 0.07\\
    TiNeuVox-B &28.17 & 0.97 & 0.07 &33.61 &0.98 & 0.03 & 31.45 & 0.97 & 0.05\\
    HexPlane &24.24 & 0.94 &0.07 &33.79 &0.98 & 0.03 &28.71 & 0.96 & 0.05\\
    \bottomrule [2 pt]
    Model &\multicolumn{3}{c}{Bouncing Balls} &\multicolumn{3}{c}{Lego} &\multicolumn{3}{c}{T-Rex}  \\
    & PSNR$\uparrow$ &SSIM$\uparrow$ &LPIPS $\uparrow$    & PSNR$\uparrow$ &SSIM$\uparrow$ &LPIPS $\uparrow$    & PSNR$\uparrow$ &SSIM$\uparrow$ &LPIPS $\uparrow$\\
    \midrule
    T-NeRF     &37.81 & 0.98 & 0.12 & 23.82  & 0.90 & 0.15 & 30.19 & 0.96 & 0.13 \\
    D-NeRF     &38.93 & 0.98 & 0.10 & 21.64  & 0.83 & 0.16 & 31.75 & 0.97 & 0.03 \\
    TiNeuVox-S &39.05 & 0.99 & 0.06 & 24.35  & 0.88 & 0.13 & 29.95 & 0.96 & 0.06 \\
    TiNeuVox-B &40.73 & 0.99 & 0.04 & 25.02  & 0.92 & 0.07 & 32.70 & 0.98 & 0.03 \\
    HexPlane  &39.69 & 0.99 & 0.03 & 25.22  & 0.94 & 0.04 & 30.67 & 0.98 & 0.03 \\
    \bottomrule [2 pt]
    Model &\multicolumn{3}{c}{Stand Up} &\multicolumn{3}{c}{Jumping Jacks} &\multicolumn{3}{c}{Average}  \\
    & PSNR$\uparrow$ &SSIM$\uparrow$ &LPIPS $\uparrow$    & PSNR$\uparrow$ &SSIM$\uparrow$ &LPIPS $\uparrow$    & PSNR$\uparrow$ &SSIM$\uparrow$ &LPIPS $\uparrow$\\
    \midrule
    T-NeRF     &31.24 & 0.97 & 0.02 & 32.01  & 0.97 & 0.03 & 29.51 & 0.95 & 0.08 \\
    D-NeRF     &32.79 & 0.98 & 0.02 & 32.80  & 0.98 & 0.03 & 30.50 & 0.95 & 0.07 \\
    TiNeuVox-S &32.89 & 0.98 & 0.03 & 32.33  & 0.97 & 0.04 & 30.75 & 0.96 & 0.07 \\
    TiNeuVox-B &35.43 & 0.99 & 0.02 & 34.23  & 0.98 & 0.03 & 32.64 & 0.97 & 0.04 \\
    HexPlane  & 34.36 & 0.98 & 0.02 & 31.65 & 0.97 & 0.04  & 31.04 & 0.97 & 0.04\\
    \bottomrule [2 pt]
    \end{tabular}}
    \label{sup:tab_dnerf}
    \vspace{-3mm}
\end{table*}

\begin{figure*}[!htp]
    \centering
    \begin{tabular}{@{}c@{\hspace{0.25mm}}c@{\hspace{0.25mm}}c@{}}
        \includegraphics[width=0.49\textwidth]{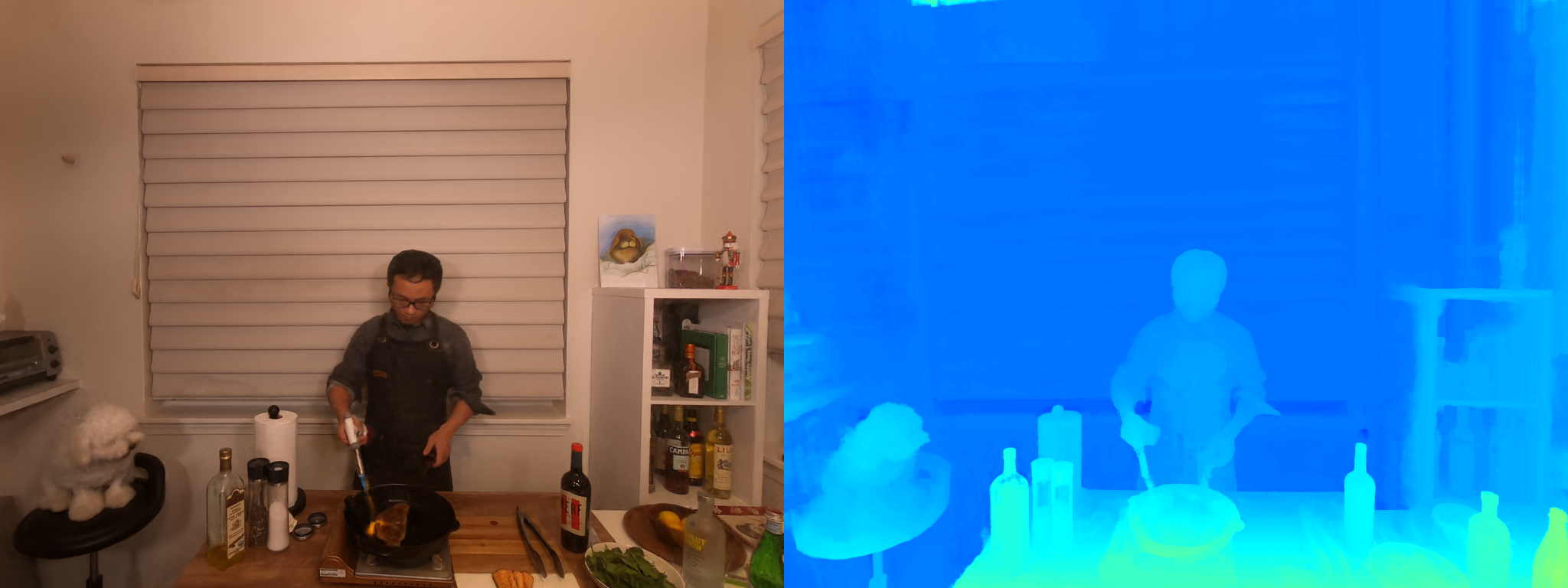} &
        \includegraphics[width=0.49\textwidth]{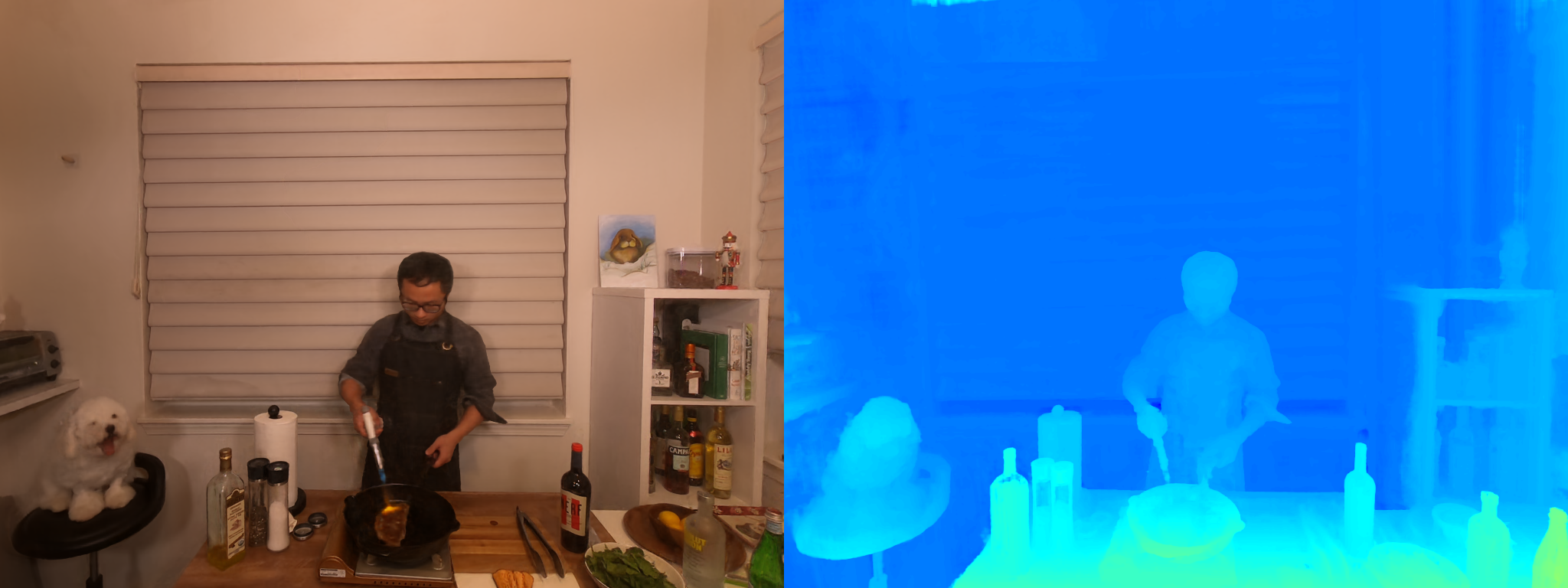} \\
        \includegraphics[width=0.49\textwidth]{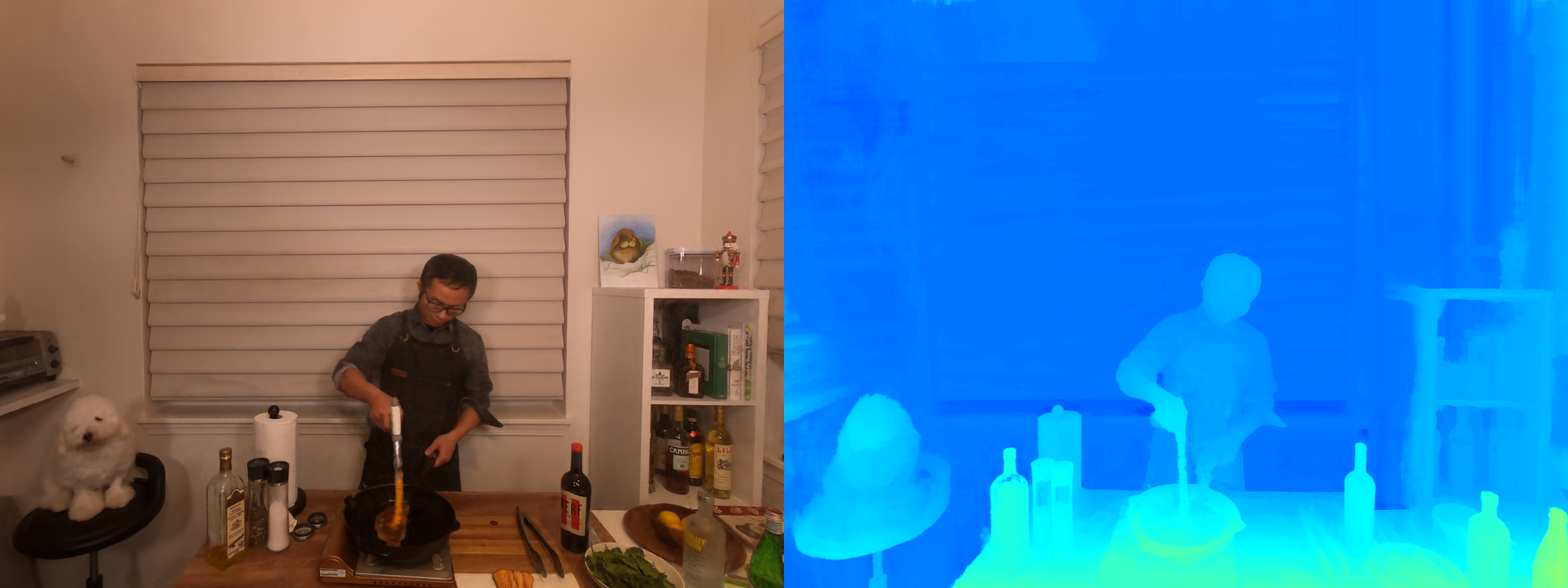} &
        \includegraphics[width=0.49\textwidth]{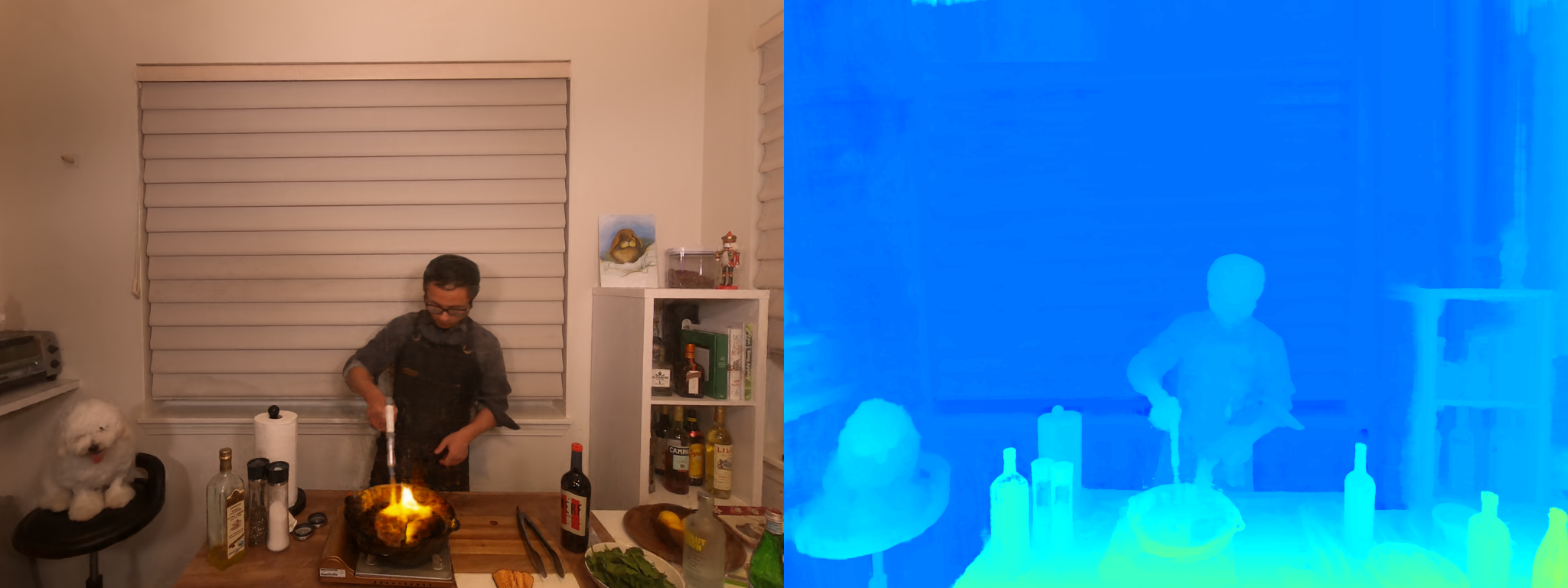} \\
        \includegraphics[width=0.49\textwidth]{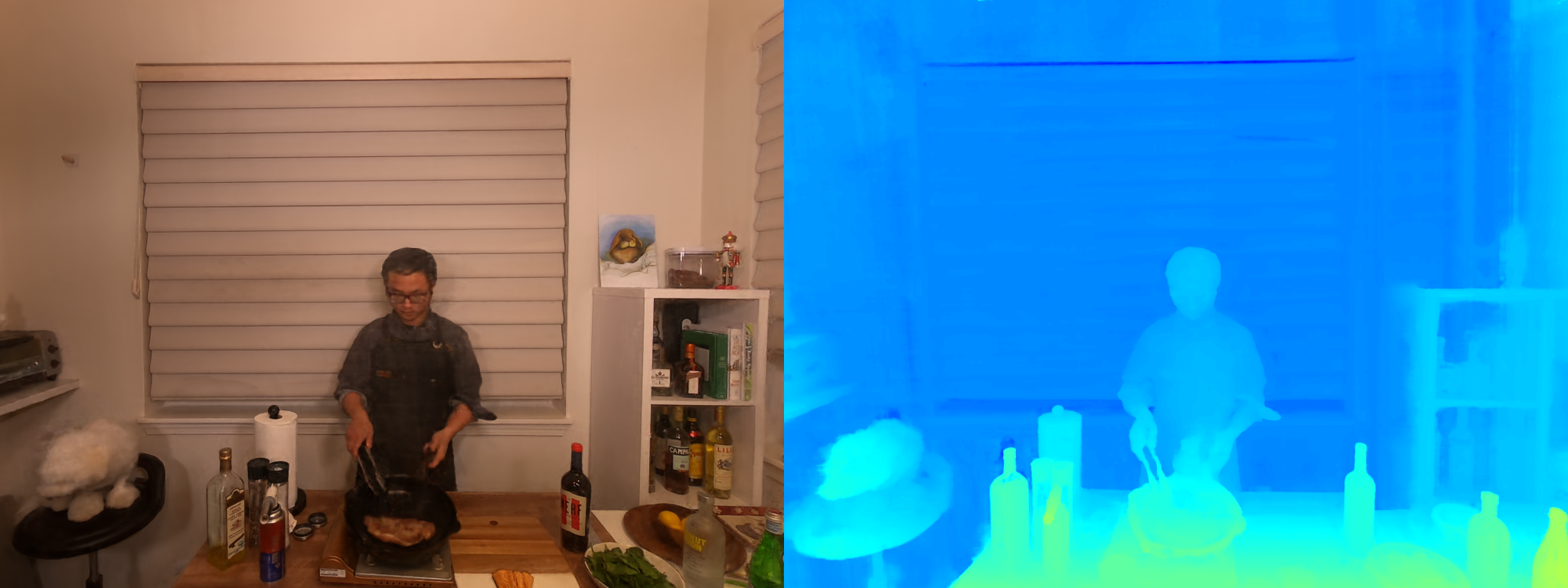} &
        \includegraphics[width=0.49\textwidth]{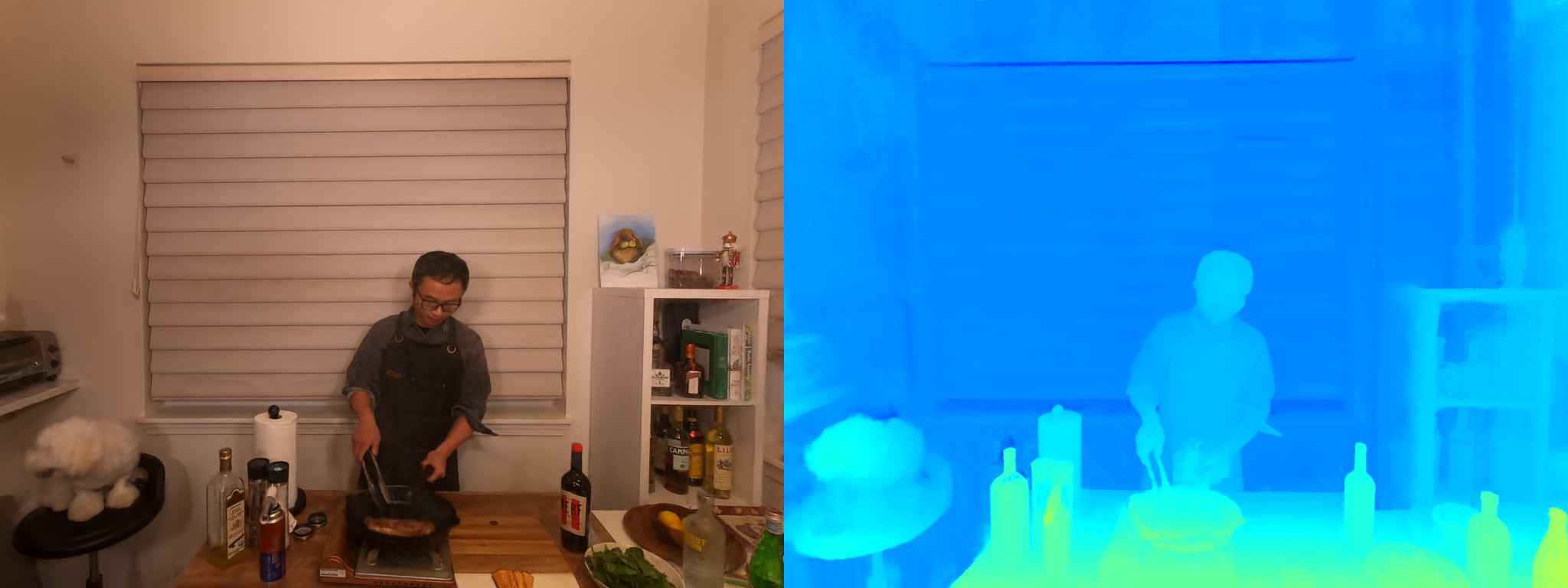} \\
        \includegraphics[width=0.49\textwidth]{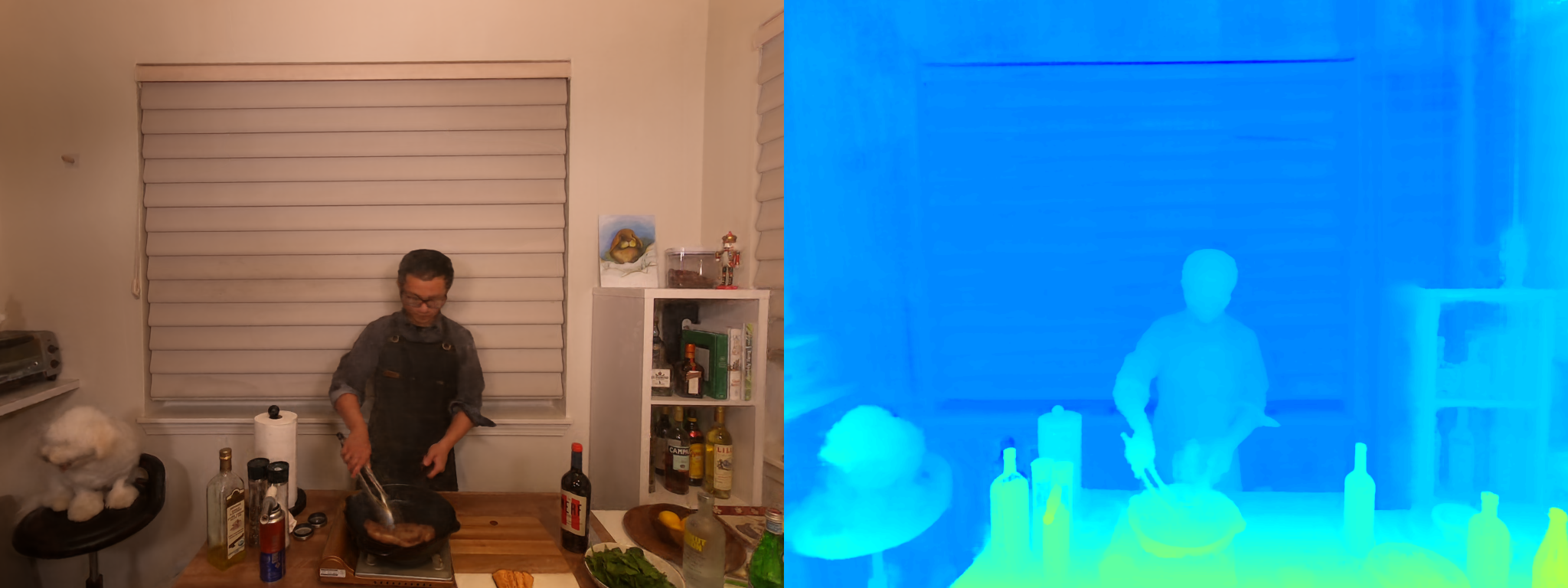} &
        \includegraphics[width=0.49\textwidth]{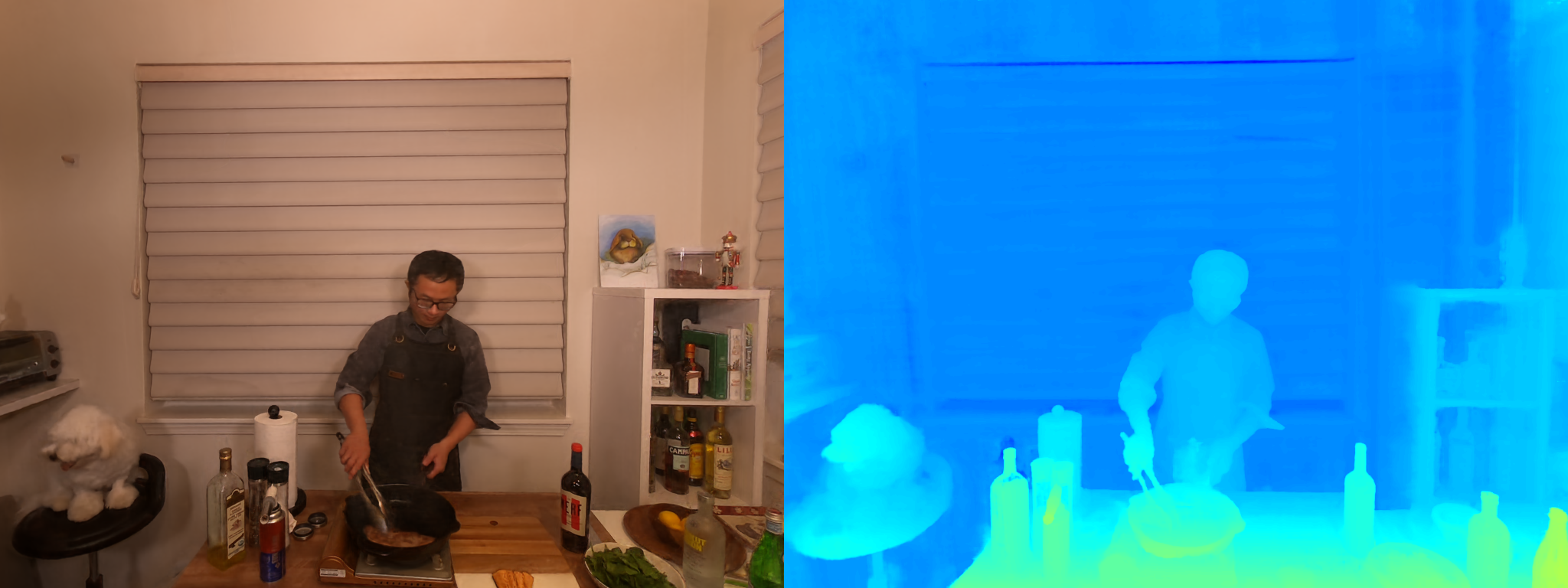} \\
        \includegraphics[width=0.49\textwidth]{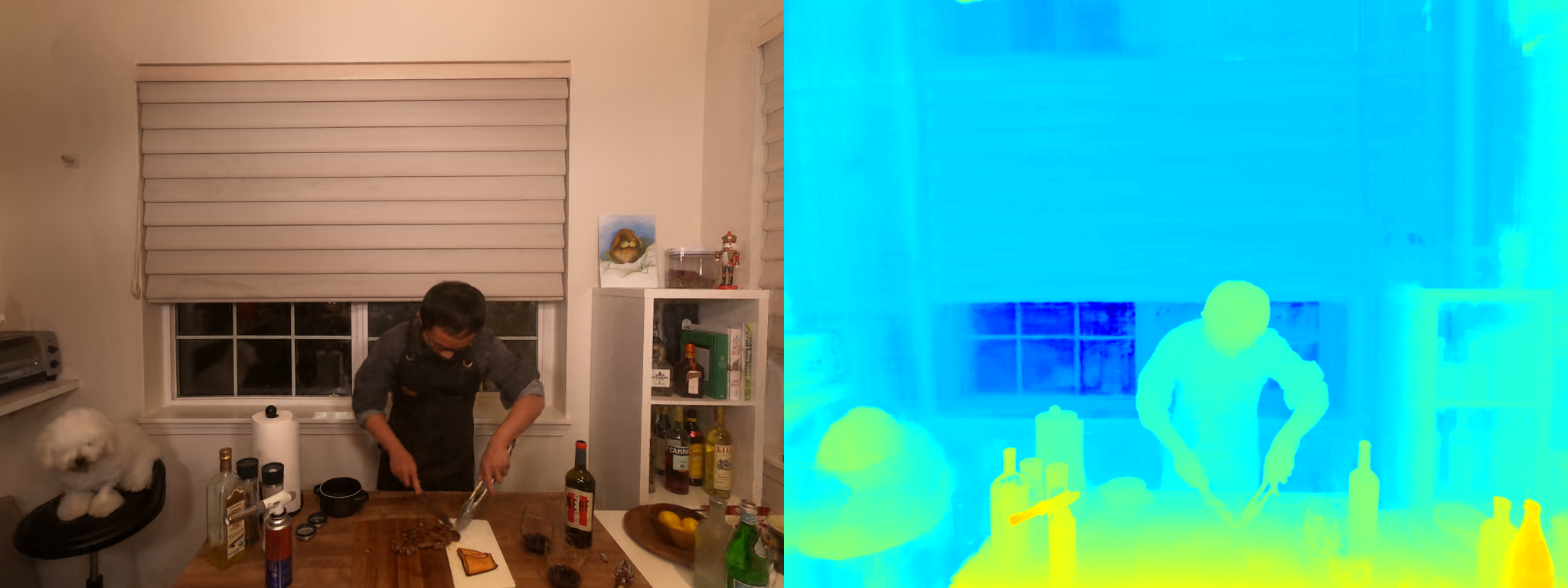} &
        \includegraphics[width=0.49\textwidth]{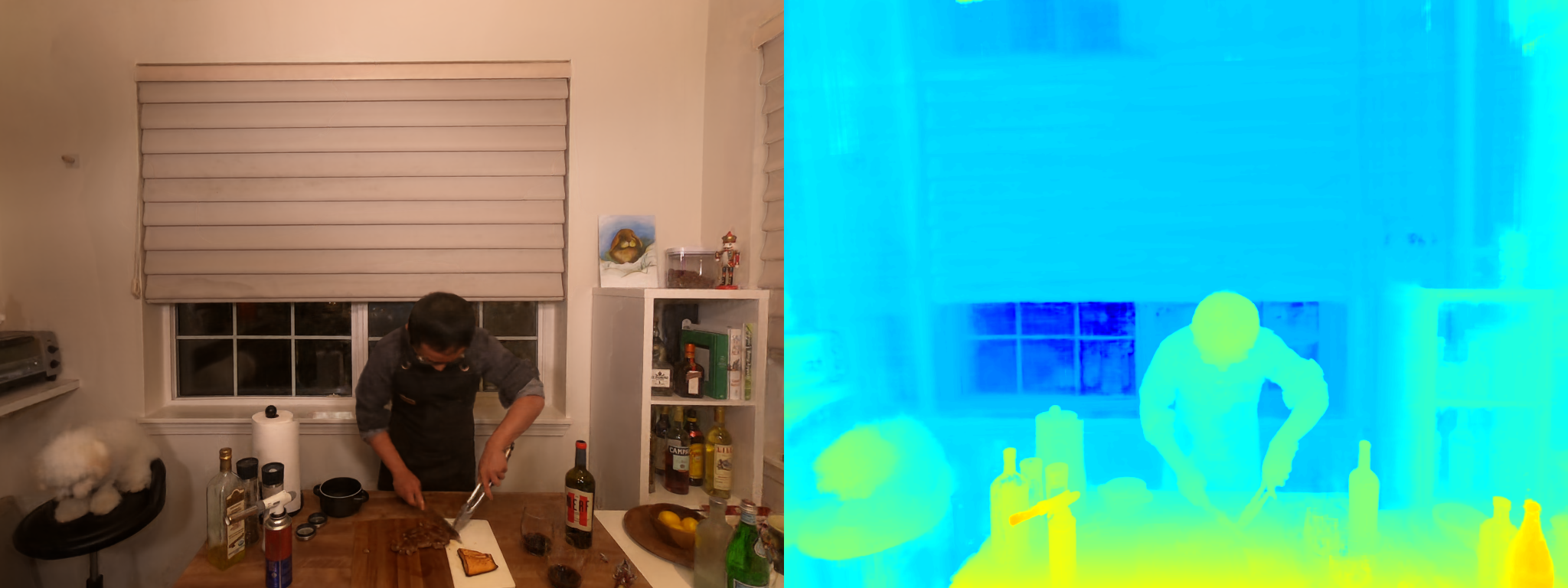} \\
        \includegraphics[width=0.49\textwidth]{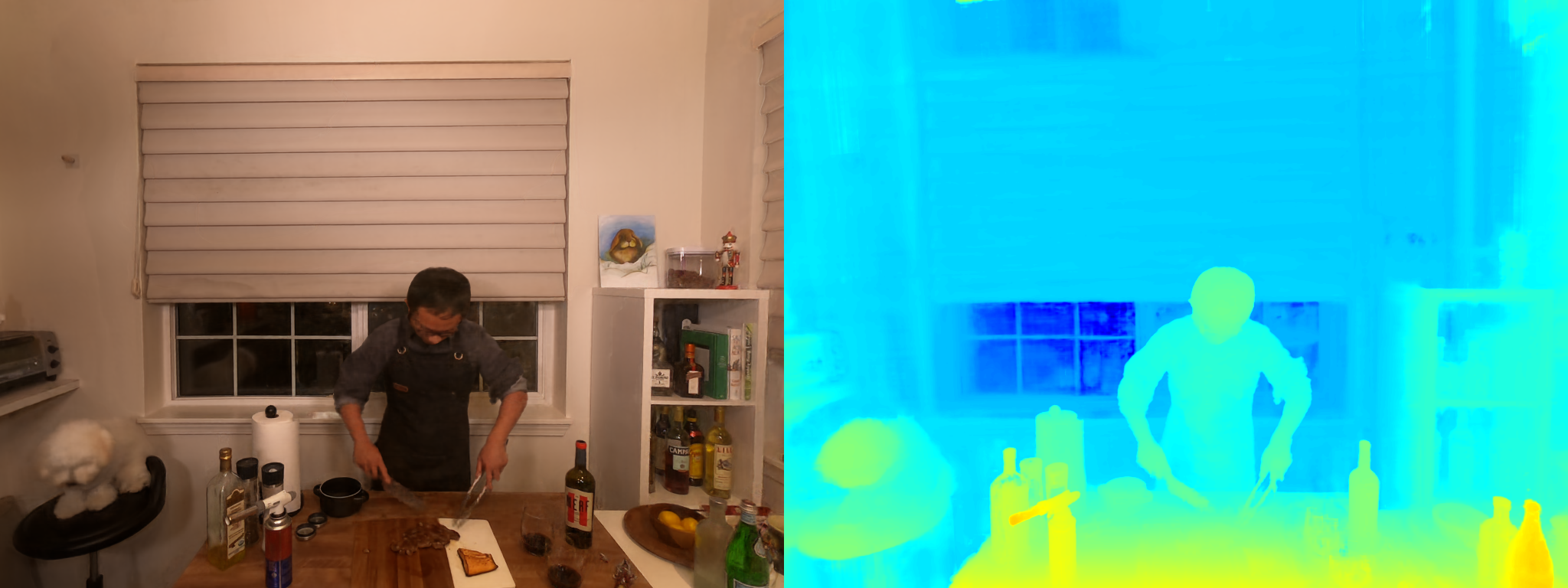} &
        \includegraphics[width=0.49\textwidth]{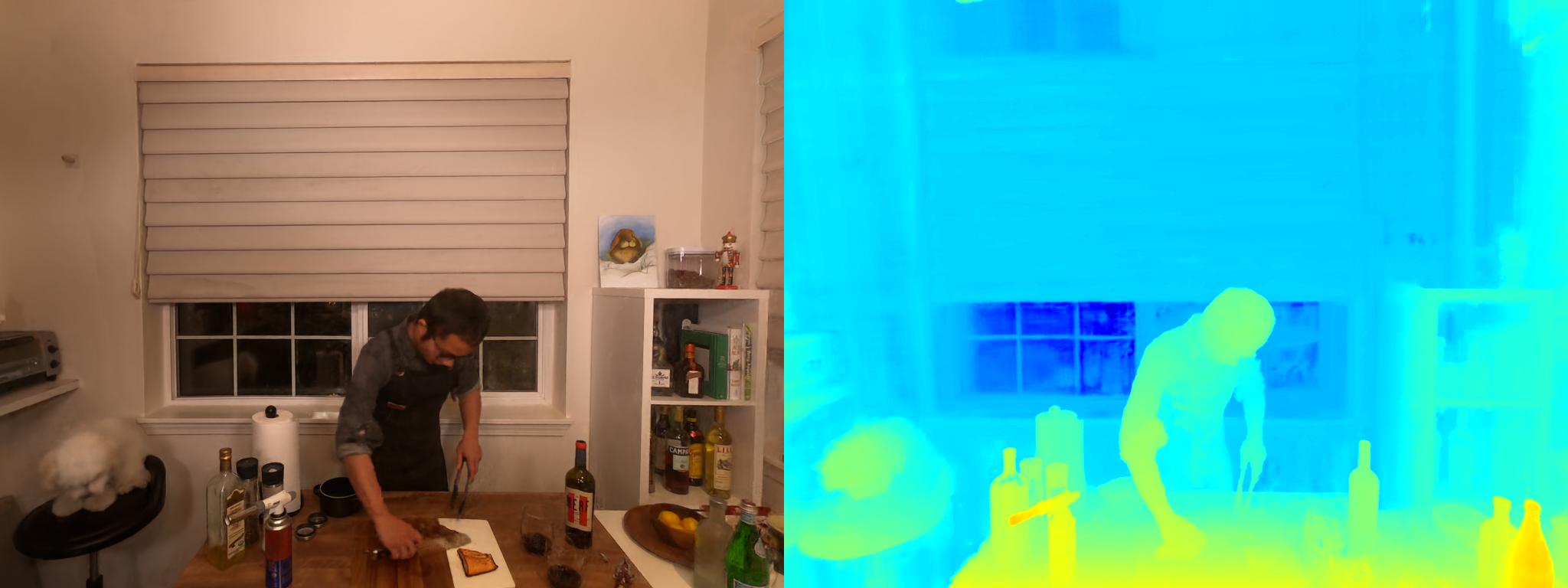} \\
    \end{tabular}
    \caption{
        \textbf{View Synthesis Results and Depths at Test View on Plenoptic Video Dataset~\cite{li2022neural}.}
        }
    \label{sup:fig_nv3d}
    \vspace{7mm}
\end{figure*}

We follow the hierarch training pipeline as ~\cite{li2022neural}.
\emph{HexPlane} in Table~1 uses 650k iterations, with 300k stage one training, 250k stage two training and 100k stage three training.
\emph{HexPlane\dag} uses 100k iterations in total, with 10k stage one training, 50k stage two training and 40k stage three training.
According to ~\cite{li2022neural}, stage one is a global-median-based weighted sampling with $\gamma = 0.001$;
stage two is also a global-median-based weighted sampling with $\gamma = 0.02$;
stage three is a temporal-difference-based weighted sampling with $\alpha=0.1$.

In evaluation, D-SSIM is computed as $\frac{1 - \text{MS-SSIM}}{ 2}$ and LPIPS~\cite{zhang2018unreasonable} is calculated using AlexNet~\cite{krizhevsky2017imagenet}.
We use default settings for Just-Objectionable-Difference (JOD)~\cite{mantiuk2021fovvideovdp}.

Each scene results are in Table~\ref{sup:tab_nv3d}, and more visualizations are in Figure~\ref{sup:fig_nv3d}.
We found that \emph{HexPlane} gives visually more smooth results than \emph{HexPlane\dag}.
Since we don't have baseline results, we don't explore new evaluation metrics. 

\subsection{D-NeRF Dataset~\cite{pumarola2021d}.}
We have $R_1 = R_2 = R_3 = 48$ for appearance HexPlane since it has $360^{\circ}$ videos.
For opacity HexPlane, we set $R_1 =R_2 =R_3 = 24$. 
The bounding box has max boundaries $[1.5, 1.5, 1.5]$ and min boundaries $[-1.5, -1.5, -1.5]$.

During training, HexPlane starts with space grid size of $32^3$ and upsamples its resolution at 3k, 6k, 9k to $200^3$.
The emptiness voxel is calculated at 4k and 10k iterations. 
Total training iteration is 25k.
The learning rate for feature planes are 0.02, and learning rate for $\mathbf{V}^{RF}$ and neural network is 0.001.
All learning rates are exponentially decayed.  
We use Adam~\cite{kingma2014adam} for optimization with $\beta_1 = 0.9, \beta_2 = 0.99$.
During evaluation, the LPIPS is computed using VGG-Net~\cite{simonyan2014very} following previous works. 
\begin{figure*}[!htp]
    \centering
    \begin{tabular}{@{}c@{\hspace{0.25mm}}c@{\hspace{0.25mm}}c@{\hspace{0.25mm}}c@{\hspace{0.25mm}}c@{}}
        \includegraphics[width=0.19\textwidth]{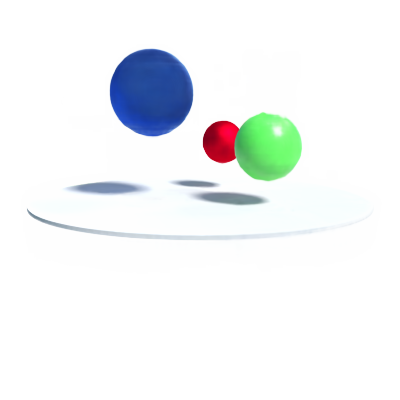} &
        \includegraphics[width=0.19\textwidth]{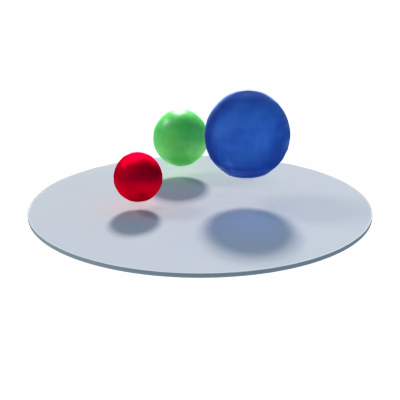} &
        \includegraphics[width=0.19\textwidth]{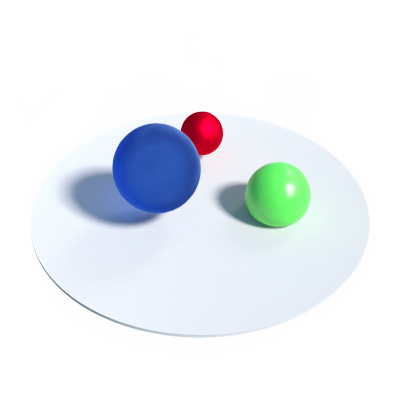} &
        \includegraphics[width=0.19\textwidth]{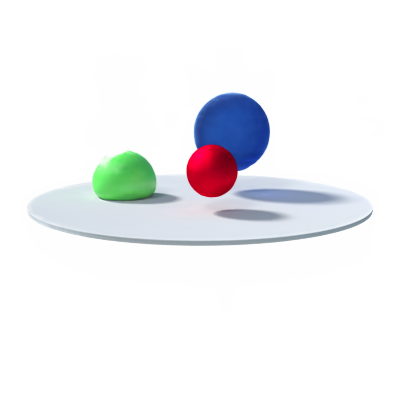} &
        \includegraphics[width=0.19\textwidth]{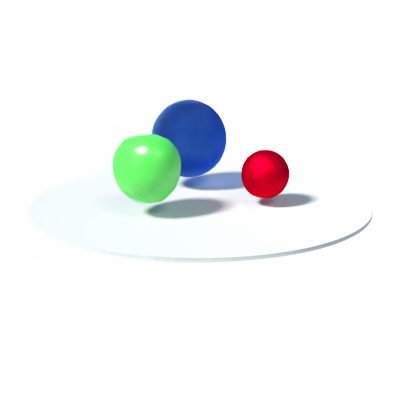} \\
        \includegraphics[width=0.19\textwidth]{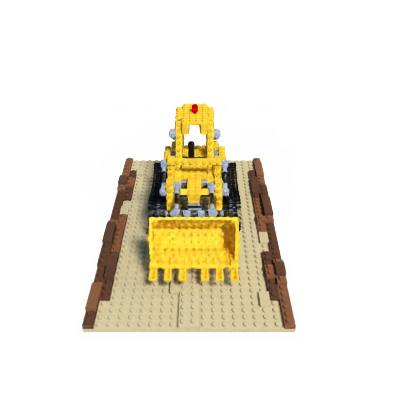} &
        \includegraphics[width=0.19\textwidth]{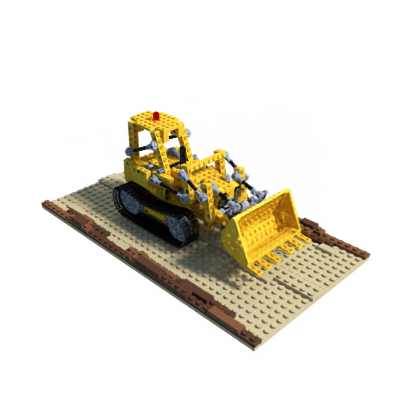} &
        \includegraphics[width=0.19\textwidth]{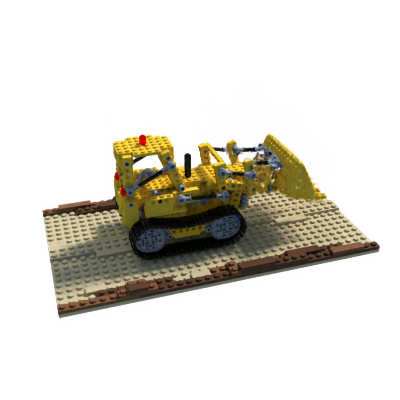} &
        \includegraphics[width=0.19\textwidth]{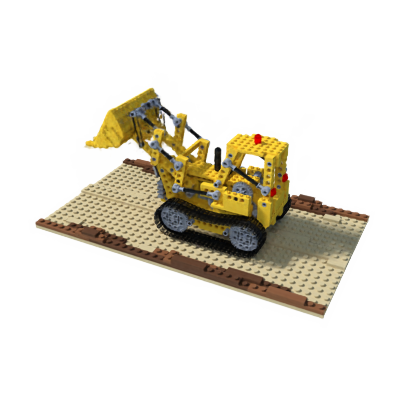} &
        \includegraphics[width=0.19\textwidth]{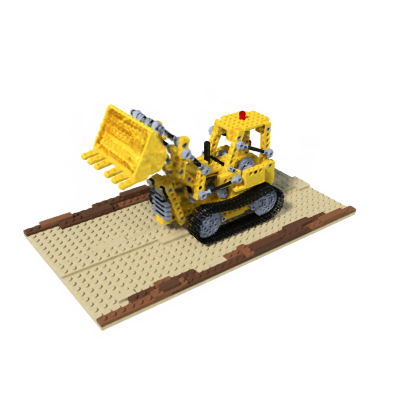} \\
        \includegraphics[width=0.19\textwidth]{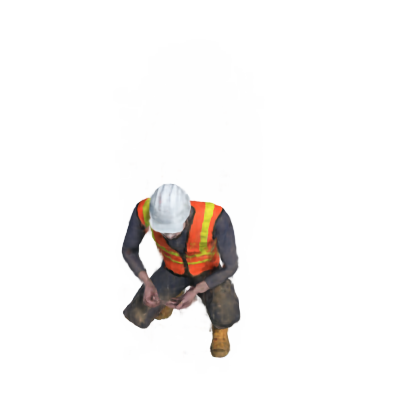} &
        \includegraphics[width=0.19\textwidth]{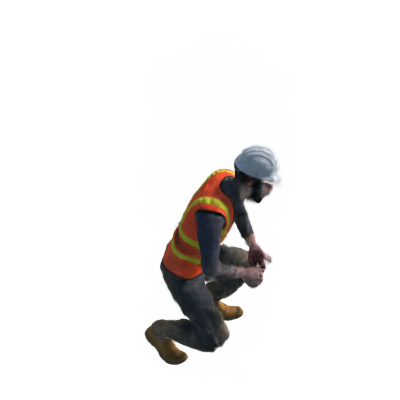} &
        \includegraphics[width=0.19\textwidth]{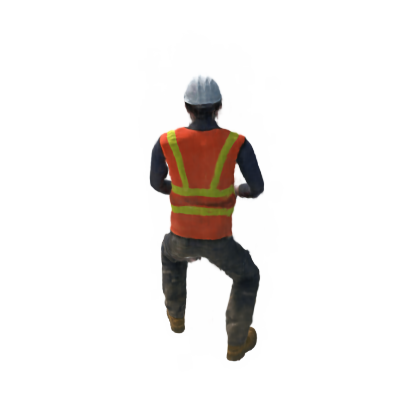} &
        \includegraphics[width=0.19\textwidth]{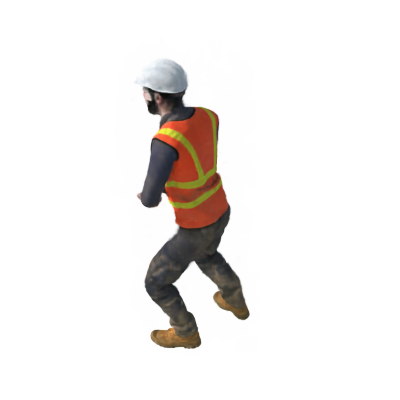} &
        \includegraphics[width=0.19\textwidth]{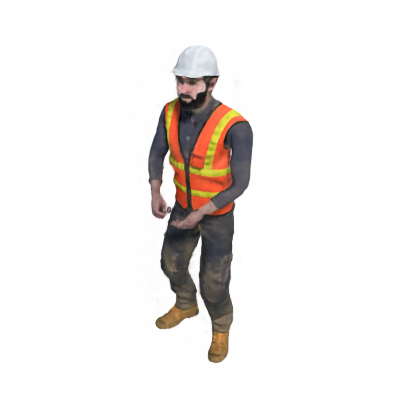} \\
        \includegraphics[width=0.19\textwidth]{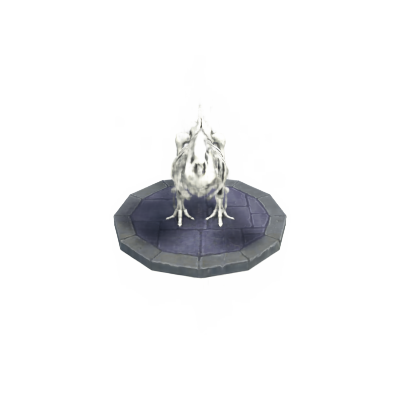} &
        \includegraphics[width=0.19\textwidth]{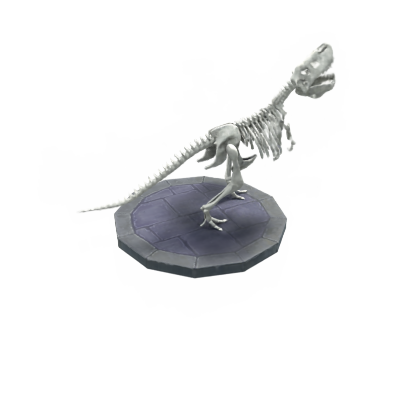} &
        \includegraphics[width=0.19\textwidth]{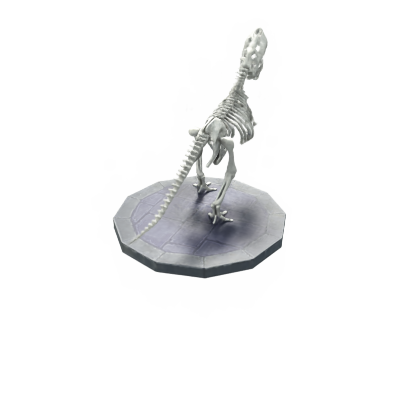} &
        \includegraphics[width=0.19\textwidth]{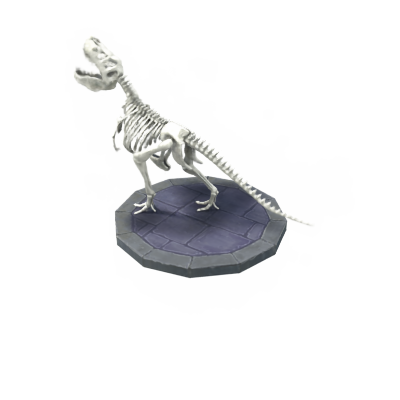} &
        \includegraphics[width=0.19\textwidth]{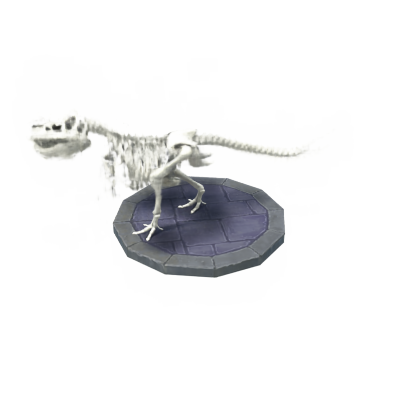} \\
        \includegraphics[width=0.19\textwidth]{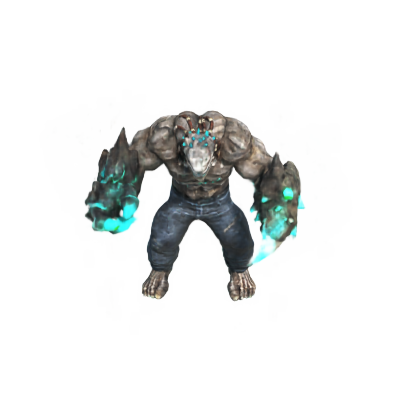} &
        \includegraphics[width=0.19\textwidth]{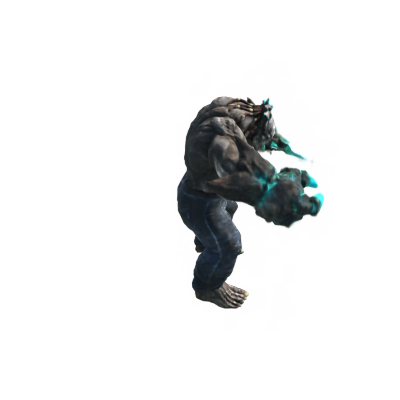} &
        \includegraphics[width=0.19\textwidth]{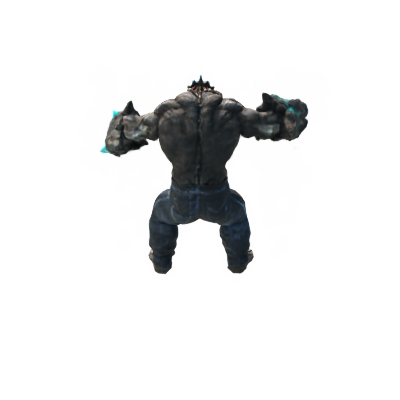} &
        \includegraphics[width=0.19\textwidth]{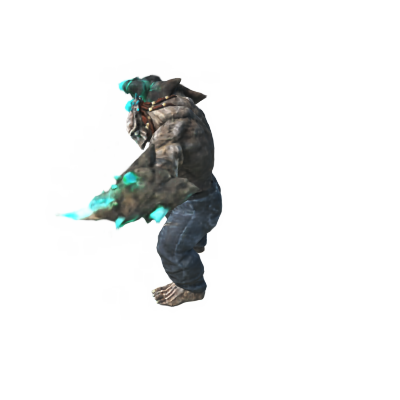} &
        \includegraphics[width=0.19\textwidth]{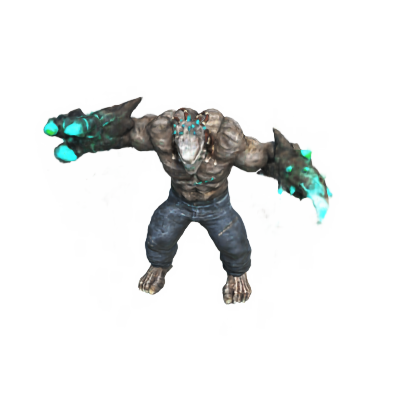} \\
        \includegraphics[width=0.19\textwidth]{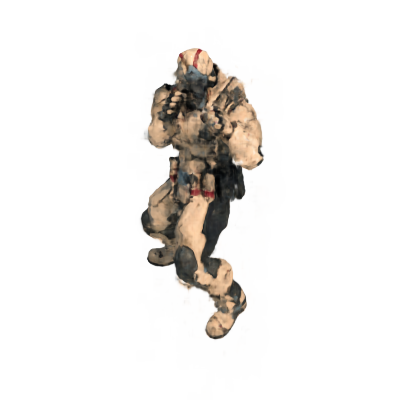} &
        \includegraphics[width=0.19\textwidth]{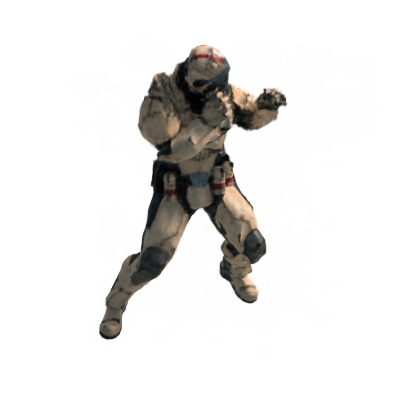} &
        \includegraphics[width=0.19\textwidth]{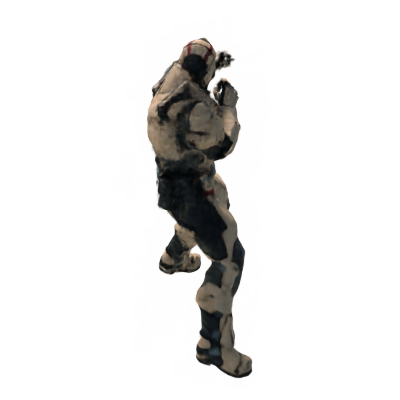} &
        \includegraphics[width=0.19\textwidth]{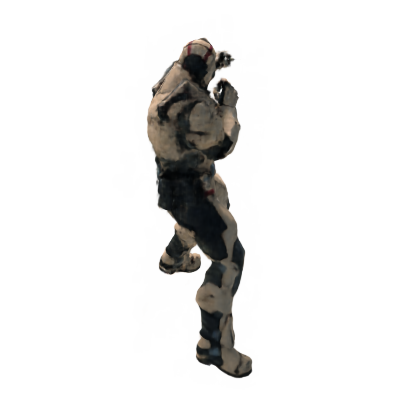} &
        \includegraphics[width=0.19\textwidth]{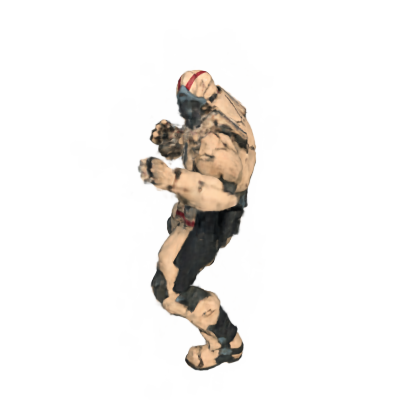} 
    \end{tabular}
    \caption{
        \textbf{View Synthesis Results on D-NeRF Dataset~\cite{pumarola2021d}.}
        }
    \label{sup:fig_dnerf}
\end{figure*}
We show per-scene quantitative results in Table~\ref{sup:tab_dnerf} and visualizations in Figure~\ref{sup:fig_dnerf}.

\subsection{iPhone dataset~\cite{gao2022dynamic}.}

\subsection{Ablation Details.}

For a fair comparison, we fix all settings in ablations.

\par \noindent \textbf{Volume Basis}
represents 4D volumes as the weighted summation of a set of shared 3D volumes as Eq 2 in main paper,
where each 3D volume is represented in Eq 1 format to save memory.
The 3D volume $\mathbf{V}_t$ at time $t$ is then:
{\footnotesize
\begin{align}
    \begin{split}
        \mathbf{V}_t = &\sum_{i=1}^{R_t} \mathbf{f}(t)_i \cdot \hat{\mathbf{V}}_i \\
        &= \sum_{i=1}^{R_t}\mathbf{f}(t)_i(\sum_{r=1}^{R_1} \mathbf{M}_{r, i}^{XY} \circ \mathbf{v}_{r,i}^Z \circ \mathbf{v}_{r, i}^1 + \sum_{r=1}^{R_2} \mathbf{M}_{r, i}^{XZ} \\
        &\circ \mathbf{v}_{r, i}^Y \circ \mathbf{v}_{r, i}^2 + \sum_{r=1}^{R_3} \mathbf{M}_{r, i}^{YZ} \circ \mathbf{v}_{r, i}^X \circ \mathbf{v}_{r, i}^3)
    \end{split}
\end{align}
}
Similarly, we use a piece-wise linear function to approximate $\mathbf{f}(t)$.
In experiments, we set $R_1=R_2=R_3=16$ for appearance HexPlane and $R_1=R_2=R_3=8$ for opacity HexPlane.
We evaluate $R_t = 8, 12, 16$ in experiments.

\par \noindent \textbf{VM-T}~(Vector, Matrix and Time) uses Eq 3 in main paper to represent 4D volumes.

\begin{align}
    \begin{split}
        \mathbf{V}_t =& \sum_{r=1}^{R_1} \mathbf{M}^{XY}_{r} \circ \mathbf{v}^{Z}_{r} \circ \mathbf{v}^{1}_{r} \cdot \mathbf{f}^1_{r}(t) + \sum_{r=1}^{R_2} \mathbf{M}^{XZ}_{r} \circ \mathbf{v}^{Y}_{r}  \\
        &\circ \mathbf{v}^{2}_{r} \cdot \mathbf{f}^2_{r}(t) + \sum_{r=1}^{R_3} \mathbf{M}^{ZY}_{r} \circ \mathbf{v}^{X}_{r} \circ \mathbf{v}^{3}_{r} \cdot \mathbf{f}^3_{r}(t)
    \end{split}
\end{align} 
We evaluate $R_1=R_2=R_3=24, 48, 96$.

\par \noindent \textbf{CP Decom.}~(CANDECOMP Decomposition) represents 4D volumes using a set of vectors for each axis.

\begin{align}
    \begin{split}
        \mathbf{V}_t =&\sum_{r=1}^{R} \mathbf{v}^{X}_{r} \circ \mathbf{v}^{Y}_{r} \circ \mathbf{v}^{Z}_{r} \circ \mathbf{v}_{r} \cdot \mathbf{f}_{r}(t)
    \end{split}
\end{align} 
$\mathbf{v}^{X}, \mathbf{v}^{Y}, \mathbf{v}^{Z}$ are feature vectors corresponding to $X, Y, Z$ axes.
We evaluate $R=48, 96, 192, 384$ in experiments. 

\subsection{Fusion Ablations}
\begin{table*}[thp!]
    \centering
    \caption{
    \textbf{Ablations on Feature Fusions Designs.}
    We show results with various fusion designs on D-NeRF dataset.
    HexPlane could work with other fusion mechanisms, showing its robustness.}
    \vspace{-3mm}
    \resizebox{0.98\textwidth}{!}{
    \begin{tabular}{p{2.5cm} p{2.5cm} P{1.5cm} P{1.5cm} P{1.5cm} P{1.5cm} P{1.5cm} P{1.5cm}}
    \toprule[2pt]
    & & \multicolumn{3}{c}{Opacity without MLP Regression} & \multicolumn{3}{c}{Opacity with MLP Regression}\\
    \cmidrule(l{2pt}r{2pt}){3-5} \cmidrule(l{2pt}r{2pt}){6-8}
    Fusion-One &Fusion-Two &PSNR$\uparrow$  & SSIM$\uparrow$ & LPIPS$\downarrow$ &PSNR$\uparrow$  & SSIM$\uparrow$ & LPIPS$\downarrow$\\
    \midrule
    \multirow{3}{*}{Multiply} & Concat & 31.042   &0.968 & 0.039 & 31.477 &0.969 &0.037 \\
    & Sum & 31.023 & 0.967 &0.039 &31.318 &0.969 &0.038 \\
    & Multiply & 30.345  & 0.966 & 0.041 & 31.094 &0.968 &0.038\\
    \midrule
    \multirow{3}{*}{Sum} & Concat  & 25.428 & 0.931 &0.084 &29.240 &0.954 & 0.057  \\
    & Sum &25.227 &0.928 &0.090 &28.024 &0.946 &0.067\\
    & Multiply& 30.585 & 0.965 &0.044 &30.934 & 0.966 & 0.041\\
    \midrule
    \multirow{3}{*}{Concat} & Concat  & 25.057 & 0.928 &0.073  &30.173 &0.961 &0.049 \\
    & Sum &24.915 &0.925 &0.077  &27.971 &0.946 &0.066\\
    & Multiply & 30.299 &0.965 &0.041 &30.874 &0.971 &0.036  \\
    \bottomrule [2 pt]
    \end{tabular}}
    \vspace{-4mm}
    \label{sup:tab_fusion_ablations}
\end{table*}

We provide complete results of fusion ablations in Table~\ref{sup:tab_fusion_ablations}.
For \emph{Fusion-One} and \emph{Fusion-Two}, we choose one fusion method from \emph{Concat}, \emph{Sum}, and \emph{Multiply}, and enumerate all combinations of fusion methods.
Besides that, we also explore to regress opacities from MLPs like~\cite{chan2022efficient}.
In this setting, we sample opacity features, 8-dim feature vectors from HexPlane and regress opacity values from another MLP.  

Using MLP to regress opacities could substantially boost the the results for all designs, at the cost of slower rendering speeds. 
Interestingly, we found that 

Please also note that we found different fusion designs expect different

\subsection{Visualization of Feature Planes.}
\begin{figure*}[!htp]
    \centering
    \includegraphics[width=0.95\textwidth]{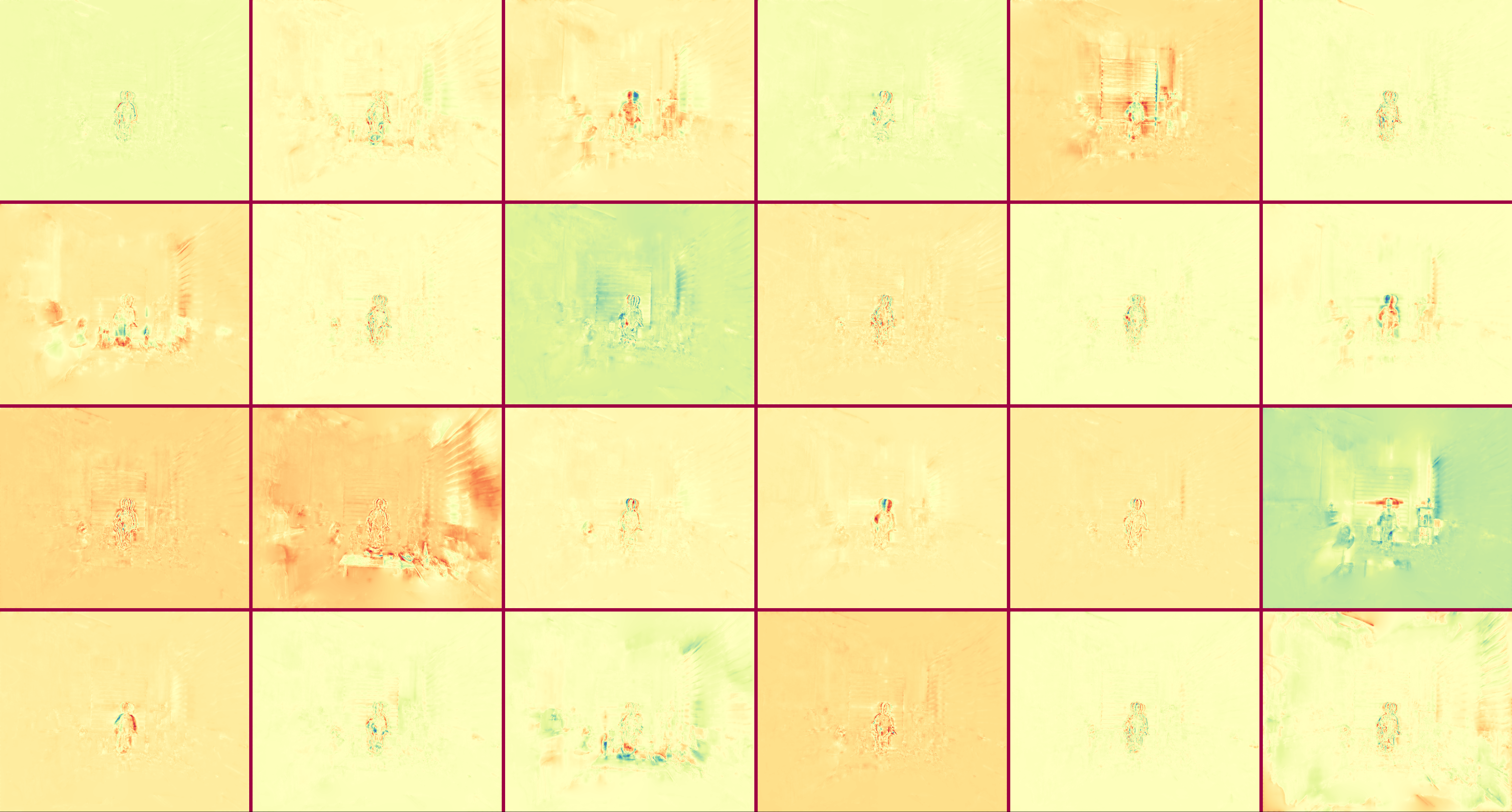}\\
    Feature Map Visualization on XY Plane \\
    \includegraphics[width=0.95\textwidth]{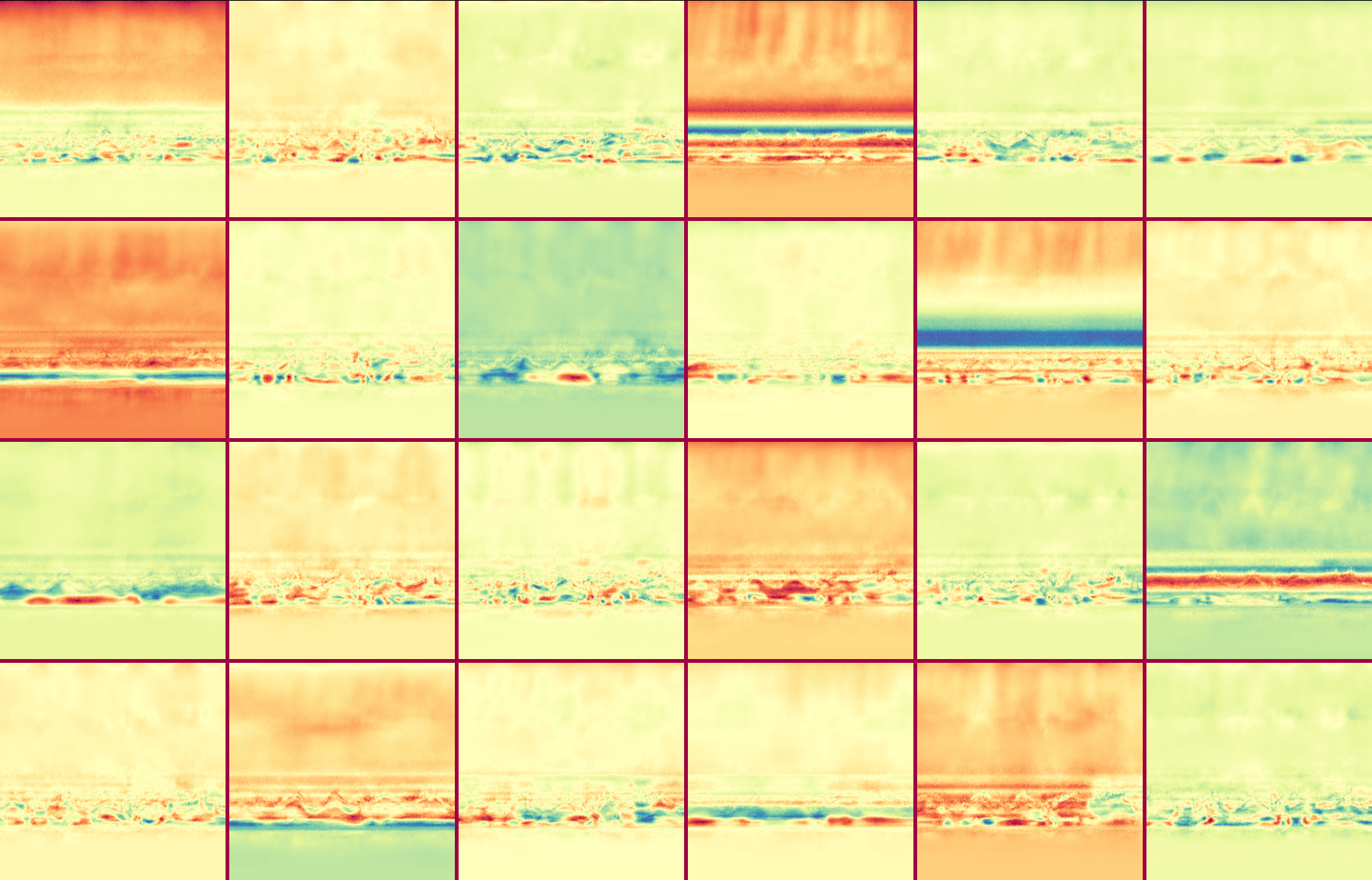}\\
    Feature Map Visualization on ZT Plane \\
    \caption{
        \textbf{Feature Map Visualization on \emph{Flame Salmon} Scene.} }
    \label{sup:fig_feat_vis}
\end{figure*}

We visualize each channel of $XT, ZT$ feature plane for opacity HexPlane in Figure~\ref{sup:fig_feat_vis}.
This HexPlane is trained in \emph{Flame Salmon} scene in Plenoptic Video Dataset~\cite{li2022neural}.

\section{Failure Cases}
\begin{figure*}[!htp]
    \centering
    \begin{tabular}{@{}c@{\hspace{0.25mm}}c@{\hspace{0.25mm}}c@{\hspace{0.25mm}}c@{}}
       Synthesis & Ground-truth & Synthesis & Ground-truth\\
        \includegraphics[width=0.24\textwidth]{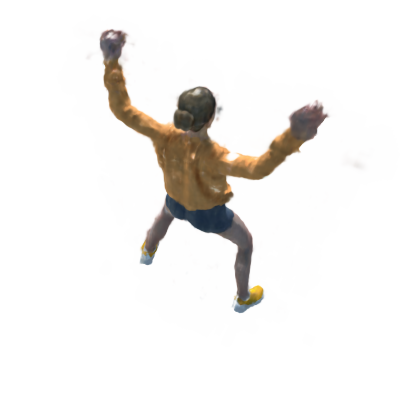} &
        \includegraphics[width=0.24\textwidth]{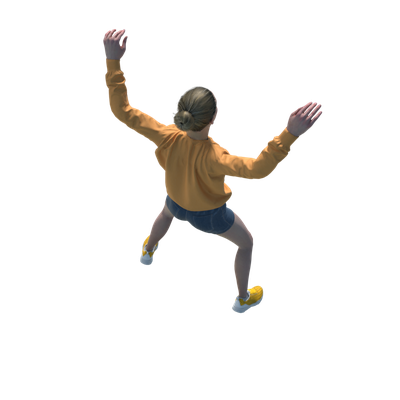} & 
        \includegraphics[width=0.24\textwidth]{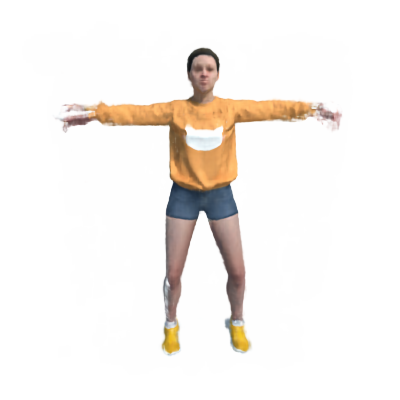} &
        \includegraphics[width=0.24\textwidth]{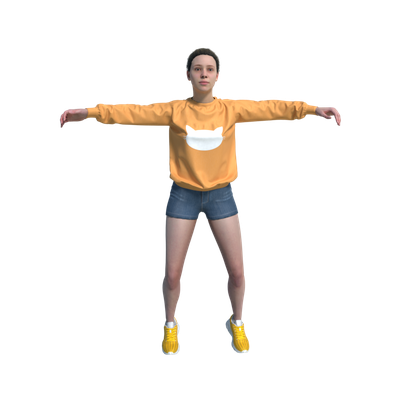}\\
        \includegraphics[width=0.24\textwidth]{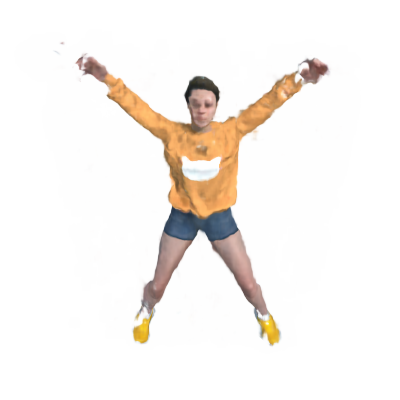} &
        \includegraphics[width=0.24\textwidth]{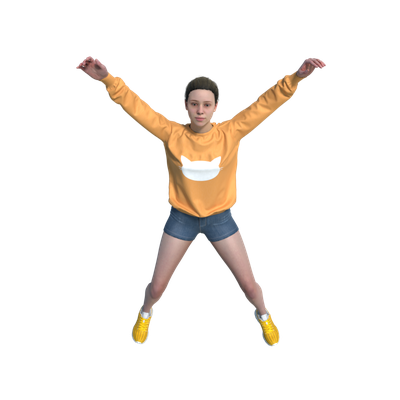} & 
        \includegraphics[width=0.24\textwidth]{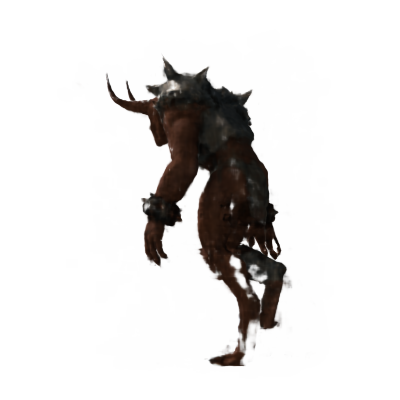} &
        \includegraphics[width=0.24\textwidth]{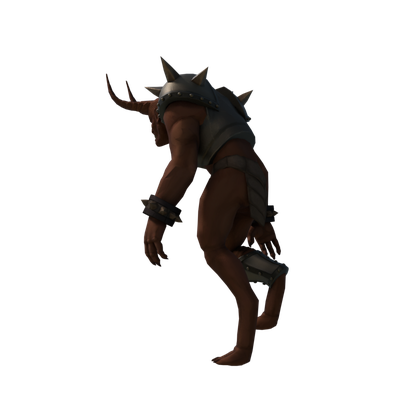}\\
        \includegraphics[width=0.24\textwidth]{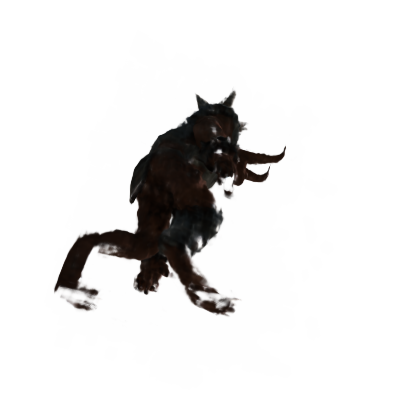} &
        \includegraphics[width=0.24\textwidth]{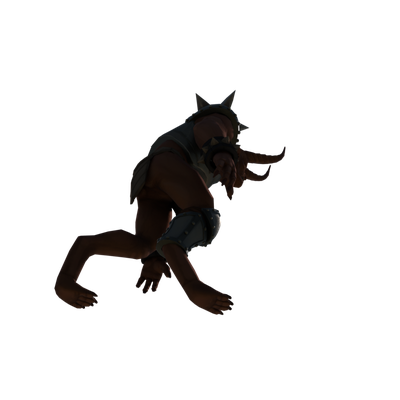} & 
        \includegraphics[width=0.24\textwidth]{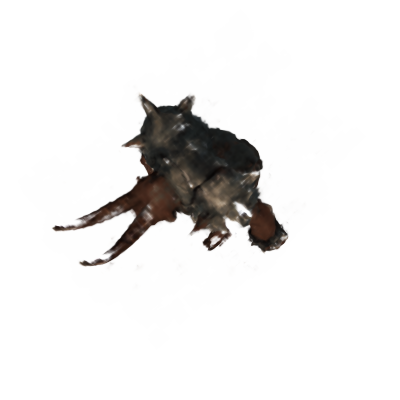} &
        \includegraphics[width=0.24\textwidth]{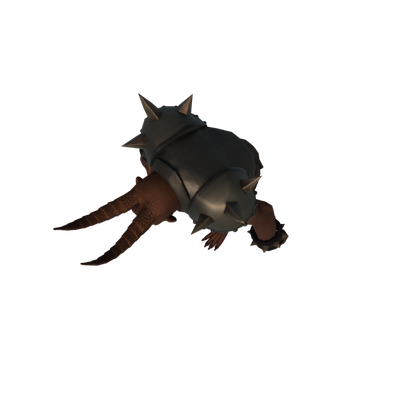}\\
    \end{tabular}
    \caption{
        \textbf{Failure Cases from HexPlane.}
        }
    \label{sup:fig_failure}
\end{figure*}
HexPlane doesn't always give satisfactory results. 
It generates degraded results when objects move too fast or there are too few observations to synthesis details.
Figure~\ref{sup:fig_failure} shows failure cases  and corresponding ground-truth images. 

\section{Failed Designs for Dynamic Scenes}
Although HexPlane is a simple and elegant solution, it is not the instant solution we had for this task.
In this section, we discuss other designs we tried. 
These designs could model the dynamic scenes while their qualities and speeds are not comparable to HexPlane.
We discuss these ``failed'' designs, hoping they could inspire future work. 

\subsection{Fixed Basis for Time Axis}
In Eq 2 of main paper, we use $\mathbf{f}(t)$ as the coefficients of basis volumes at time $t$.
Its could be further expressed as:
\begin{equation}
    \mathbf{V_t} = \sum_{i=1}^{R_t} \mathbf{f}(t)_i \cdot \hat{\mathbf{V}}_{i} = \hat{\mathbf{V}} \cdot \mathbf{f}(t)
\end{equation}
where the second $\cdot$ is matrix-vector production; $\hat{\mathbf{V}} \in \mathbb{R}^{XYZFR_t}$ is the stack of $\{\hat{\mathbf{V}}_{1}, \ldots, \hat{\mathbf{V}}_{R_t}\}$;
$\mathbf{f}(t) \in \mathbb{R}^{R_t}$ is a function of $t$.
An interesting perspective to understand $\hat{\mathbf{V}} $ is: instead of storing static features with shape $\mathbb{R}^F$, every spatial point in 3D volume contains a feature matrix with shape $\mathbb{R}^{FR_t}$.
And feature vectors at specific time $t$ could be computed by inner product between $\mathbf{f}(t)$ and feature matrix.
That is, $\mathbf{f}(t)$ is a set of basis functions w.r.t to time $t$ and feature matrix contains coefficients of basis functions to approximate feature value changes along with time. 
Following the traditional approach of basis functions for time series, we use a set of sine/cosine functions as $\mathbf{f}(t)$.

While in practice, we found this implementation couldn't work since it requires enormous GPU memories. 
For instance, with $X=Y=Z=128, R_t=32, F=27$, it uses 7GB to store such a representation and around 30GB during training because of back-propagation and keeping auxiliary terms of Adam.
And it is extremely slow because of reading/writing values in memories.

Therefore, we apply tensor decomposition to reduce memory usages by factorizing volumes into matrixes $\mathbf{M}^{XY},\mathbf{M}^{XZ},\mathbf{M}^{YZ}$  and vectors $\mathbf{v}^{X}, \mathbf{v}^{Y},\mathbf{v}^{Z}$ following Eq 1.
Similarly, we add additional $R_t$ dimension to matrixes $\mathbf{M}$ and vectors $\mathbf{v}$, leading to $\mathbf{MT}_{r}^{XY}\in \mathbb{R}^{XYR_T}, \mathbf{vT}^{X}_{r} \in \mathbb{R}^{XR_T}$.
When calculating features from the representation, $\mathbf{f}(t)$ is first multiplied with $\mathbf{MT}_{i}^{XY}$ and $\mathbf{vT}^{X}_{i}$ along the last dimension to get  $\mathbf{M}$ and $\mathbf{v}$ at this time steps.
We then calculate features using resulted $\mathbf{v}$ and  $\mathbf{M}$ following Eq 1. 

$\mathbf{f}(t)$ is designed to be like positional encoding, $\mathbf{f}(t) = [1, \sin(t), \cos(t), \sin(2*t), \cos(2*t), \sin(4*t), \cos(4*t), \cdots]$.
We also try to use Legendre polynomials or Chebyshev polynomials to represent $\mathbf{f}(t)$.
During training, we use weighted $L1$ loss to regularize $\mathbf{MT}_{r}, \mathbf{vT}_{r}$, and assign higher weights for high-frequency coefficients to keep results smooth.
We also use the smoothly bandwidth annealing trick in~\cite{park2021nerfies} during training, which gradually introducing high-frequency components.

This design could model dynamic scenes while it suffers from severe color jittering and distortions. 
Compared to HexPlane, it has additional matrix-vector production, which reduces overall speeds.
\subsection{Frequency-Domain Methods}
We also tried another method from the frequency domain, which is orthogonal to the HexPlane idea. 
The notations of this section are slightly inconsistent with the notations of the main paper. 

According to Fourier Theory, the value at $(x, y, z, t)$ spacetime point could be represented in its frequency domain (we ignore feature dimension here for simplicity):
\begin{align}
    \begin{split}
        \mathbf{D}(x,y,z,t) = &\sum_{u=1}^{U} \sum_{v=1}^{V} \sum_{w=1}^{W} \sum_{k=1}^{K}  \\
        &\widetilde{\mathbf{D}}(u, v, w, k) \cdot e^{-j2\pi (\frac{ux}{U} + \frac{vy}{V} + \frac{wz}{W} + \frac{kt}{K})}
    \end{split}
\end{align}
$\widetilde{\mathbf{D}}$ is another 4D volume storing frequency weights, having the same size as $\mathbf{D}$.
Storing $\widetilde{\mathbf{D}}$  is memory-consuming, and similarly, we apply tensor decomposition on this volume.
{\footnotesize
\begin{align}
    \begin{split}
        & \mathbf{D}(x,y,z,t) = \sum_{u=1}^{U} \sum_{v=1}^{V} \sum_{w=1}^{W} \sum_{k=1}^{K} \sum_{r=1}^{R} \widetilde{\mathbf{v}}^{U}(u)_r \cdot \widetilde{\mathbf{v}}^{V}(v)_r \cdot \\
        & \widetilde{\mathbf{v}}^{W}(w)_r \cdot \widetilde{\mathbf{v}}^{K}(k)_r \cdot e^{-j2\pi (\frac{ux}{U} + \frac{vy}{V} + \frac{wz}{W} + \frac{kt}{K})} \\
     = &\sum_{u=1}^{U} \sum_{v=1}^{V} \sum_{w=1}^{W} \sum_{k=1}^{K} \sum_{r=1}^{R} (\widetilde{\mathbf{v}}^{U}(u)_r \cdot e^{-j2\pi \frac{ux}{U}}) (\widetilde{\mathbf{v}}^{V}(v)_r \cdot e^{-j2\pi \frac{vy}{V}}) \\
      &  \cdot (\widetilde{\mathbf{v}}^{W}(w)_r e^{-j2\pi \frac{wz}{W}}) \cdot (\widetilde{\mathbf{v}}^{K}(k)_r e^{-j2\pi \frac{kt}{K}})) \\
    = & \sum_{r=1}^{R}(\sum_{u=1}^{U}\widetilde{\mathbf{v}}^{U}(u)_r \cdot e^{-j2\pi \frac{ux}{U}}) \cdot \sum_{v=1}^{V}(\widetilde{\mathbf{v}}^{V}(v)_r \cdot e^{-j2\pi \frac{vy}{Y}}) \cdot \\
    &(\sum_{w=1}^{W}\widetilde{\mathbf{v}}^{W}(w)_r e^{-j2\pi \frac{wz}{W}}) \cdot (\sum_{k=1}^{K}\widetilde{\mathbf{v}}^{K}(k)_r e^{-j2\pi \frac{kt}{K}})
    \end{split}
\end{align}
}
where $\widetilde{\mathbf{v}}^{U},\widetilde{\mathbf{v}}^{V},\widetilde{\mathbf{v}}^{W},\widetilde{\mathbf{v}}^{K}$ are  the decomposed vectors from $\widetilde{\mathbf{D}}$ along $U, V, W, K$ axes using CANDECOMP Decomposition,
which axes are related to $x, y, z, t$ in time domain. 
Instead of storing the 4D frequency volume and computing values by traversing all elements inside this volume using Eq 5, 
we decompose 4D volumes into many single vectors and calculate values by summation along each axis, significantly reducing computations.

Similarly, we apply weight $L1$ and smoothly bandwidth annealing trick on vector weights. 
We also try wavelet series instead of Fourier series, and other decompositions. 
We found this method leads to less-saturated colors and degraded details, which is shown in videos.

Also, this method replaces grid sampling of HexPlane by inner product, which is less efficient and leads to slow speeds. 
\end{document}